\documentclass{elsarticle}

\usepackage[english]{babel}
\usepackage[utf8x]{inputenc}
\usepackage[T1]{fontenc}
\usepackage{ dsfont }
\usepackage{placeins}
\usepackage{multirow}
\usepackage{lscape}
\usepackage{algpseudocode}
\usepackage{color,soul}
\usepackage{hyperref} 
\usepackage{csvsimple}
\usepackage{longtable}
\usepackage{arydshln,amssymb,color}
\usepackage{multicol}
\usepackage{geometry}
\usepackage{lscape}
\usepackage{amsmath, amsthm, amssymb}
\usepackage{xparse,etoolbox}

\newtheorem*{prop}{Proposition}
\newcommand{\blocktheorem}[1]{%
	\csletcs{old#1}{#1}
	\csletcs{endold#1}{end#1}
	\RenewDocumentEnvironment{#1}{o}
	{\par\addvspace{1.5ex}
		\noindent\begin{minipage}{\textwidth}
			\IfNoValueTF{##1}
			{\csuse{old#1}}
			{\csuse{old#1}[##1]}}
		{\csuse{endold#1}
		\end{minipage}
		\par\addvspace{1.5ex}}
}
\blocktheorem{prop}

\usepackage{subcaption}

\usepackage{amsmath}
\usepackage{graphicx}
\usepackage[colorinlistoftodos]{todonotes}
\usepackage{booktabs} 
\usepackage{xtab,booktabs}
\usepackage{bm}
\usepackage{mwe}

\usepackage[linesnumbered,algoruled,boxed,lined]{algorithm2e}
\SetKwInOut{KwParam}{Parameters}
\DeclareMathOperator*{\abs}{abs}

\DeclareMathOperator*{\Matrix}{\mathcal{M}}
\newcommand{\bigO}{\mathcal{O}}

\usepackage{booktabs} 

\journal{Swarm and Evolutionary Computation}

\begin{document}
	
	\begin{frontmatter}

		\title{Kernels of Mallows Models under the Hamming Distance for solving the Quadratic Assignment Problem}

		\author[BCAM]{Etor Arza}
		\ead{earza@bcamath.org}
		\author[BCAM]{Aritz Pérez}
		\ead{aperez@bcamath.org}
		\author[BCAM]{Ekhiñe Irurozki}
		\ead{eirurozki@bcamath.org}
		\author[EHU]{Josu Ceberio}
		\ead{josu.ceberio@ehu.eus}
		
		\address[BCAM]{BCAM - Basque Center for Applied Mathematics, Spain}
		\address[EHU]{University of the Basque Country UPV/EHU, Spain}

		\begin{abstract}
        The Quadratic Assignment Problem (QAP) is a well-known permutation-based combinatorial optimization problem with real applications in industrial and logistics environments. 
        Motivated by the challenge that this NP-hard problem represents, it has captured the attention of the optimization community for decades.
        As a result, a large number of algorithms have been proposed to tackle this problem.
		Among these, exact methods are only able to solve instances of size $n<40$.
		To overcome this limitation, many metaheuristic methods have been applied to the QAP.

		In this work, we follow this direction by approaching the QAP through Estimation of Distribution Algorithms (EDAs). 
		Particularly, a non-parametric distance-based exponential probabilistic model is used. 
		Based on the analysis of the characteristics of the QAP, and previous work in the area, we introduce Kernels of Mallows Model under the Hamming distance to the context of EDAs. 
		Conducted experiments point out that the performance of the proposed algorithm in the QAP is superior to (i) the classical EDAs adapted to deal with the QAP, and also (ii) to the specific EDAs proposed in the literature to deal with permutation problems.

		\end{abstract}

	\end{frontmatter}

	\section{Introduction}

	The Quadratic Assignment Problem (QAP) \cite{QAPcreated} is a well-known combinatorial optimization problem. 
	Along with other problems, such as the traveling salesman problem, the linear ordering problem and the flowshop scheduling problem, it belongs to the family of permutation-based (a permutation is a bijection of the set $\{1,...,n\}$ onto itself) problems \cite{ceberio2014solving}.
	The QAP has been applied in many different environments over the years, to name but a few notable examples, selecting optimal hospital layouts \cite{QAPhospital}, optimally placing components on circuit boards \cite{QAPcircuit}, assigning gates at airports \cite{QAPairport} or optimizing data transmission \cite{mittelmann_solving_2015}.

	\begin{algorithm}
		
		\DontPrintSemicolon 
		\caption{Estimation of Distribution Algorithm \cite{ceberio2014solving}}
		\label{algo:barebones_EDA}
		\KwParam{ \ \\ 
			\noindent \underline{$P_s$}: The size of the population used by the EDA. \\
			\noindent \underline{M}: The size of the set of selected solutions. \\
			\noindent \underline{S}: The number of new solutions generated at each iteration. \\ \ \\
		}

		$D_0$ $\gets$ initialize population of size $P_s$ and evaluate the population\; 
		\For{t=1,2,... until stopping criterion is met}{
			$D_{t-1}^{sel}$ select $M \leq N$ from $D_{t-1}$ according to a selection criterion\;
			$p_t(x) = p(x | D_{t-1}^{sol}) \gets$ estimate a probability distribution from $D_{t-1}^{sol}$\;
			$D_t^S \gets$ sample S individuals from $p_t(x)$ and evaluate the new individuals\; 
			$D_t \gets$ create a new population of size $P_s$ from $D_{t-1}$ and $D_t^S$\; 
		} 
	\end{algorithm}

	Sahni and Gonzalez \cite{QAPnphard} proved that the QAP is an NP-hard optimization problem, and as such, no polynomial-time exact algorithm can solve this problem unless P=NP.
	In this sense, until recently, only a few instances of size up to 36 were solved using exact solution algorithms.
	In fact, and exceptionally, only some instances of size 64 and 128 have been solved by using a combination of three strategies:
	reformulation to a suitable Mixed-Integer Linear Programming, exploiting the sparsity and symmetry of some particular instances, and a Branch and Bound algorithm (B\&B) \cite{LARGEBB4_RECENT}.
	These strategies, however, require the instance to be highly symmetric, and in general, this cannot be guaranteed.
	A quick literature review shows that the vast majority of exact methods for the QAP are based on the B\&B algorithm \cite{LARGEBB2}.
	In order to overcome the high computation cost required by these algorithms, Astreicher et al. \cite{LARGEBB1} proposed a grid computing implementation of B\&B.
	Under this technique, this algorithm can be distributed over the internet, forming a computational grid, with the advantage of bringing down the costs and increasing the availability of parallel computation power.

	Unfortunately, despite the previous improvements, in the general case, it is still computationally unfeasible to use exact algorithms for medium and large size instances $(n > 40)$. 
	In response to this drawback, the community of combinatorial optimization has proposed a variety of metaheuristic algorithms to tackle the QAP. 
	A few of these proposed methods include Genetic Algorithms~\cite{GA_2,GA_3}, Tabu Search~\cite{TABU_QAP1,CPTS}, Simulated Annealing~\cite{QAP_SA2}, Ant Colony Optimization~\cite{ANT2}, Memetic Algorithms~\cite{MEMETIC1} and Particle Swarm Optimization Algorithms~\cite{PSO_1_QAP}.

	Currently, three of the best performing approaches for the QAP are Cooperative Parallel Tabu Search (CPTS)~\cite{james_cooperative_2009-1}, a Memetic Search algorithm (MS)~\cite{MEMETIC2} and the Parallel Hyper-Heuristic on the Grid (PHHG)~\cite{dokeroglu_novel_2016}. In particular, CPTS is based on the successful robust tabu search implementation~\cite{taillard1991robust}.
	This tabu search implementation is simpler and is executed in parallel with a shared pool of solutions, aiming to promote both the quality and the diversity of the solutions available to each tabu search execution.
	Memetic search algorithms combine population based algorithms with local search procedures.
	The MS implementation~\cite{MEMETIC2} uses a uniform crossover operator, a breakout local search procedure (a local search procedure with perturbation mechanics to overcome local optima)~\cite{benlic_breakout_2013}, a fitness based replacement strategy, and an adaptive mutation strategy.
	Lastly, PHHG executes different metaheuristic algorithms in parallel, and based on the performance of those algorithms, repeatedly executes the most successful metaheuristics.
	Even though these three methods are very different from each other, they all share a property: they are highly complex hybrid procedures.

	Not limited to the previous approaches, Estimation of Distribution Algorithms (EDAs)~\cite{EDAintro1} have also been used to solve the QAP.
	Zhang et al. presented an hybrid approach of guided local search and a first marginal based EDA for the QAP~\cite{zhang_combination_2003}. 
	Pradeepmon et al.~\cite{pradeepmon_hybrid_2018} applied a hybrid EDA to the keyboard layout problem, which is modeled as a QAP problem.
	Previous EDA references use hybridization techniques to improve the performance of the algorithms, nevertheless, the probability models used are simplistic and are not suited to the characteristics of the QAP.
	In this paper, we investigate probability models specially suited to deal with the QAP, and introduce a new EDA proposal based on these models.

	An Estimation of Distribution Algorithm (EDA) \cite{EDAintro1} is a population-based evolutionary algorithm (see Algorithm \ref{algo:barebones_EDA} for the general pseudo-code of EDAs).
	Starting with an initial population (line 1), a subset of the best solutions is selected (line 2).
	Subsequently, a probability model is learned based on these selected permutations (line 4).
	Next, new solutions are sampled from the probability model, and their objective value is computed (line 5).
	Finally, the new solutions are combined with the selected solutions to create a new population (line 6).
	This process is repeated until a stopping criterion is met, such as exceeding a certain time constraint or a maximum number of iterations.

	%
	%
	%
	%
	%

	As reported frequently in the literature, the behavior of an EDA depends highly on the probability model used in the learn-sample cycle, as this is the core component of the algorithm.
	When considering permutation problems, EDAs can be classified into three categories according to the probability domain of the model used \cite{ceberio2012review}.
	In the first group, we have EDAs that were designed to solve problems on the combinatorial domain.
	Specifically, EDAs that were designed for the set $\{(k_1,...,k_n) \in \mathbb{N}^n : k_i\leq n\ \forall i\}$ (denoted as $[n]^n$ in this work) can be adapted for the permutation space.
	This adaptation is based on the idea that the solution space of the QAP (the set of every permutation of size n, $\mathcal{S}^n$) is a subset of $[n]^n$.
	Therefore, given a set of solutions from $\mathcal{S}^n$, a probability model can be learned in $[n]^n$.
	Consequently, in order to obtain solutions that are in $\mathcal{S}^n$, the sampling procedure must be adapted to guarantee that the new samples are in $\mathcal{S}^n$.

	In the second group, we have EDAs that were originally designed to deal with continuous domain problems.
	Next, in order to deal with permutations, these EDAs use a mapping $\gamma: \mathbb{R}^n \longrightarrow \mathcal{S}^n$, such that, given a real vector $v \in \mathbb{R}^n$, $\gamma(v)$ denotes the order of the items in $v$.
	EDAs for continuous problems transform each permutation $\sigma_i$ in the population into a real vector $v_i$, making sure $\gamma(v_i) = \sigma_i$ is satisfied for all the individuals in the population.
	Then, a probability model is learned on $\mathbb{R}^n$ from these real vectors, and new real vectors are sampled from this probability model.
	Finally, $\gamma$ is applied to all the new real vectors, to obtain the new solutions.
	One of the major drawbacks of these models, as stated by Bosman et al. \cite{rk_overhead}, are the overheads introduced by the sorting of the real vectors, which can be costly in certain situations.
	Another limitation of the models in this group comes from the large redundancy introduced by the codification, since a permutation can be mapped by infinite real vectors \cite{ceberio2012review}.

	Finally, in the third group, we have the EDAs that use probability models that define a probability distribution on $\mathcal{S}^n$.
	Among these, we find probability models based on order statistics, such as the Plackett-Luce~\cite{plackett1975analysis} and Bradley-Terry~\cite{hunter2004} models, or those that rely on distance-based distributions on $\mathcal{S}^n$ such as the Mallows Model (MM)~\cite{MMcreation} and the generalized Mallows Model (GMM)~\cite{GMM}.
	For these EDAs, the solution space of the problem is also the space onto which the probability distribution is defined, making them a more natural choice.
	The MM is an exponential distribution that can be seen as analogous to the normal distribution over the group $\mathcal{S}^n$.	
	Recently, it has been shown that the distance-metric under which an MM is defined influences the performance of the EDA \cite{review_on_distances}.
	In this sense, most of the MMs presented in the literature are based on the Cayley, Kendall's-$\tau$ and Ulam distances.
	For example, Ceberio et al. have applied many variants of MM based EDAs to different permutation problems, including Cayley based Kernels of MM~\cite{ceberio_kernels_2015} and Generalized MM~\cite{GMMcayleyQAP} on the QAP.
	In addition, a MM variant for the bi-and tri-objective FSP was proposed by Zangari et al.~\cite{zangari2018decomposition}.
	In fact, Ceberio et al. obtained state-of-the-art results in the FSP by hybridizing a MM EDA with a variable neighborhood search~\cite{ceberio2014distance}.
	This illustrates the importance of developing efficient and adequate probability models.

	In addition to the three metrics mentioned above, there are other distance-metrics on $\mathcal{S}^n$ that have not previously been considered in EDAs, such as the Hamming distance-metric \cite{irurozki2014r, GMM_HAMMING}.
	The Hamming distance between two permutations counts the number of point-wise disagreements and is a natural choice for measuring the distance between assignments or matchings.

	In this paper, extending the work in~\cite{arza_approaching_2019}, we take a step forward in the development of EDAs specific for assignment type permutation problems by using probabilistic models based on the Hamming distance. 
	Specifically, the proposed probabilistic model is a Mallows model-based kernel density that uses the Hamming distance.
	The goal of this paper is not to present a state-of-the-art algorithm.
	Instead, a methodological contribution is made to the design of probabilistic models of EDAs for permutations.
	Particularly, we show why the probability model used in the introduced EDA is suitable for the QAP, one of the most popular assignment type permutation problems. 
	In addition, by studying the properties of the QAP, and based on the experimentation results, we claim that the Hamming-based kernel density probability model proposed in this paper is suited to be used in EDAs for solving assignment problems in the solution space of permutations of a certain size\footnote{In order to take a step forward in the design of EDAs and avoid misinterpretations, in this paper, we will only consider EDAs in their base form (no hybridization).}.


	Another relevant feature of the MM is that it is a unimodal model, and is centered at a given central permutation. 
	The unimodality and symmetry properties imposed by the MM can be too restrictive in certain contexts, not allowing multimodal scenarios to be accurately modelled~\cite{MMbadKMMgood}. 
	However, the MM can be suitable as a building block in more complex models.
	An alternative that breaks these strong assumptions is the kernel density estimate using Mallows kernels (KMMs). 
	Instead of having a central permutation, KMMs spread the probability mass by using a non-parametric averaging of MMs centered at each solution. 
	This allows the distribution to model probability distributions more accurately over the space of permutations when the strong assumptions of the MM are not fulfilled by the set of solutions.

	Taking advantage of this flexibility, the EDA approach presented in this manuscript implements a KMM under the Hamming distance.
	For the sake of analyzing the performance of the proposed algorithm, we conduct three experiments.
	First, we compare the proposed approach to other Hamming-based MM approaches.
	Then, we see how it compares to other EDAs that use probability models specific to $\mathcal{S}^n$.
	Finally, we show that Hamming KMM EDA is better than other classical EDAs.
	Specifically, conducted experiments show that the proposed approach is better than other EDAs in the literature in terms of lower Average Relative Deviation Percentage (ARDP).
	Moreover, the use of both Kernels and Hamming seems to be necessary for the best possible performance.

	The rest of the paper is organized as follows: in the following section, we briefly explain the QAP and the adequacy of the Hamming distance for this problem.
	Next, in Section \ref{section:mm}, we introduce the kernels of Mallows Models over the Hamming distance.
	Then, in Section \ref{section:eda}, we detail the proposed algorithm.
	Afterwards, in Section \ref{section:experiments} we present the experimentation, and Section \ref{section:conclusion} concludes the article.

	\section{The Quadratic Assignment Problem and the Hamming distance}
	\label{section:QAP}

	The Quadratic Assignment Problem (QAP) is the problem of optimally allocating $n$ facilities at $n$ locations in order to minimize a cost function related to the flow and the distance between every pair of facilities and pair of locations.
	In the QAP, an instance is defined by two matrices $D$, $H$ $\in \Matrix\limits_{n \times n} (\mathds{R^+})$, where $D_{i,j}$ is the distance between locations $i$ and $j$, and $H_{l,k}$ is the flow between facilities $l$ and $k$.
	The search space of the QAP is the set of every permutation $\sigma$ of size $n$, $\mathcal{S}^n$.
	In this sense, in the QAP, the aim is to find the permutation $\sigma \in \mathcal{S}^n$ that describes the optimal assignment of facilities into locations, where $\sigma(i) = j$ denotes that the $j^{th}$ facility is assigned to the $i^{th}$ location.

	When Koopmans et al.~\cite{QAPcreated} introduced the QAP, they presented it as a maximization problem.
	Given two cost matrices $D$, $H$ $\in \Matrix\limits_{n \times n}$ and a profit matrix $A$ $\in \Matrix\limits_{n \times n}$, the QAP was formulated as shown in Equation~\eqref{equation:FOriginal}.

	\begin{equation}
	\label{equation:FOriginal}		
	\max\limits_{\sigma \in \mathcal{S}^n} \sum_{i=1}^{n} A_{\sigma(i), \sigma(j)} - \sum_{i=1}^{n} \sum_{j=1}^{n} D_{i,j} H_{\sigma(i), \sigma(j)}
	\end{equation} \vspace{0.05cm}
	
	Later on, the lineal term $\sum_{i=1}^{n} A_{\sigma(i), \sigma(j)}$ was dropped, since the complexity of the problem lies within the quadratic term $\sum_{j=1}^{n} D_{i,j} H_{\sigma(i), \sigma(j)}$, \cite{loiola_survey_2007-1}.

	\begin{equation}
	\label{equation:FitnessFunction}		
	\min\limits_{\sigma \in \mathcal{S}^n}  \sum_{i=1}^{n} \sum_{j=1}^{n} D_{i,j} H_{\sigma(i), \sigma(j)}
	\end{equation} \vspace{0.05cm}

	There is an equivalent formulation that is more popular among exact methods.
	With this notation, the QAP is formulated as an integer problem:

	\begin{equation}
	\label{equation:FBB}		
	\min  \sum_{i,j=1}^{n} \sum_{k,p=1}^{n} D_{i,j} H_{k,p} x_{i,j} x_{k,p}
	\end{equation}
	restricted to
	\begin{equation*}
	\sum_{j=1}^{n} x_{i,j} = \sum_{i=1}^{n} x_{i,j} = 1, \ \ \ \forall i,j \in \{1,2,...,n\}
	\end{equation*} 
	
	\begin{equation*}
	x_{i,j} \in \{0,1\}, \ \ \ \forall i,j \in \{1,2,...,n\}
	\end{equation*}

	The parameter term $D_{i,j} H_{k,p}$ can be rewritten as $C_{i,j,p,q} = D_{i,j} H_{k,p}$.
	By doing so, a generalization of the QAP was proposed \cite{lawler1963quadratic}.
	In this generalization, the parameters $C_{i,j,p,q}$ are not a product of two values.
	Instead, each $C_{i,j,p,q}$ is considered as an independent parameter of a problem instance.

	Other more complex variants have been proposed~\cite{loiola_survey_2007-1}, such as the multiobjective QAP~\cite{knowles2002towards}.
	However, in this paper, we only focus on the currently most popular permutation based QAP variant, see Equation~\eqref{equation:FitnessFunction}.
	We argue that this version of the QAP is the most widely studied version of the QAP.

	\subsection{Distance-metrics}
	\label{sect:distanceMetrics}

	The MM relies on the definition of the distance for permutations, and three have been primarily considered in the framework of EDAs: Cayley, Kendall's-$\tau$ and Ulam~\cite{review_on_distances,ceberio2014distance,ceberio2014solving}.
	The Cayley distance measures the minimum number of swaps needed to transform a permutation into another one.
	The Kendall's-$\tau$ distance measures the number of differently arranged pairs of items between two permutations.
	Finally, the Ulam distance between permutations $\sigma$ and $\pi$ is equal to the size of the permutations, $n$, minus the length of the longest increasing subsequence in $\sigma \pi^{-1}$.
	In addition to the previous distance-metrics, the Hamming metric also been reported for the case of permutations.
	The Hamming distance between two permutations, $\sigma$ and $\pi$, counts the number of point-wise disagreements they have.

\begin{figure}
		\centering
		\includegraphics[width=0.48\textwidth]{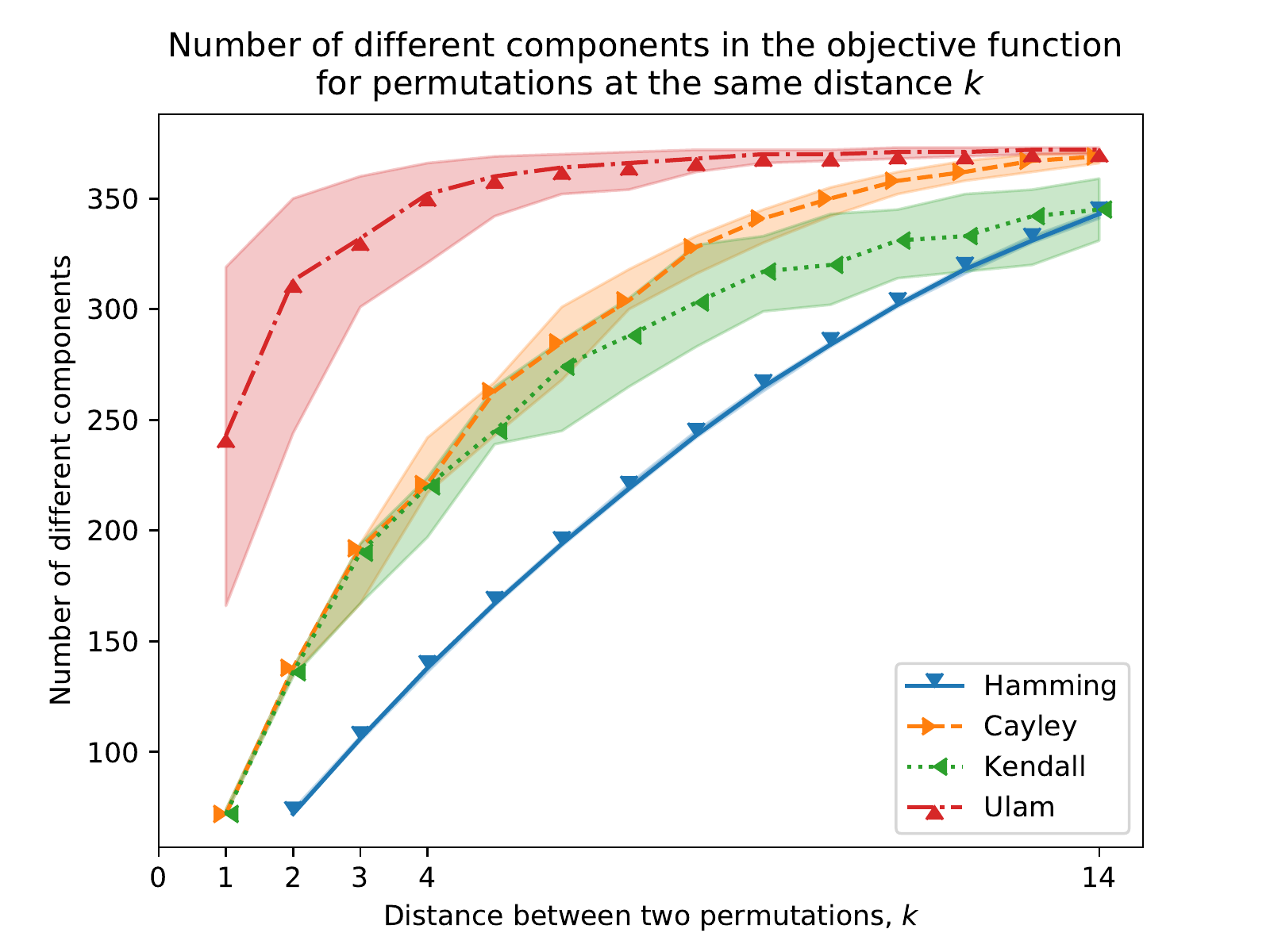}
		\caption{
			The median, $25\%$ and $75\%$ percentiles of the maximum number of different components that two solutions at a given distance can have in an instance of size $n=20$.
			The Hamming distance has the lowest number of different components among the four distance-metrics studied.
			In addition, the Hamming distance has the most consistent (almost constant) number of different components at a given distance, followed by the Cayley and Kendall distances.
			The Ulam distance is the worst distance in terms of number of different components.
		}		
		\label{fig:dist_params}
	\end{figure}

	As mentioned in the introduction, the distance employed in the MM critically conditions the performance of the EDA when solving a given problem.
	Previous work on this topic \cite{review_on_distances, fitness_distance_correlation} demonstrated that it is crucial to choose operators (distances, neighborhoods, mutations,...) that better fit the characteristics of the problem.
	We carried out an experiment in order to analyze the correlation between the distance at which two permutations are and the number of components that differ in both permutations, where component refers to each additive term $D_{i,j} H_{\sigma(i), \sigma(j)}$ for $i,j\in[n]$ in Equation~\eqref{equation:FitnessFunction}.	
	Intuitively, it is preferable when a distance-metric structures solutions in the way that close solutions differ in few components and far away solutions differ greatly.
	In this way, two solutions that are similar in terms of distance will likely be similar in terms of objective function components.
	The experiment consists of the following: For each of the considered four metrics, we choose two permutations at distance $k$ from each other. 	
	Then, we measured the maximum number of different components they can have on a problem of size $n=20$, independently of the instance.
	Finally, we repeated this process several times with different permutations, at the same distance $k$, in order to obtain the median and the interquartile range of the values.	
	The results are depicted in Figure~\ref{fig:dist_params}.

	As can be observed, the Hamming distance-metric shows the best results among the considered distance-metrics.
	On the one hand, it presents the least number of different components at each distance $k$.
	On the other hand, the number of different components is the same for all the permutations at distance $k$ (unlike the rest of the metrics).

\begin{figure}
		\begin{subfigure}{0.48\textwidth}
			\includegraphics[width=\linewidth]{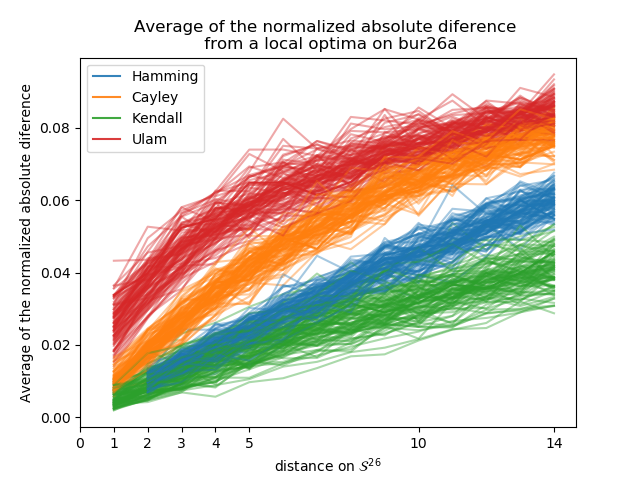} 
			\caption{bur26a}
			\label{fig:aSmoothness}
		\end{subfigure}
		\hspace*{\fill} 
		\begin{subfigure}{0.48\textwidth}
			\includegraphics[width=\linewidth]{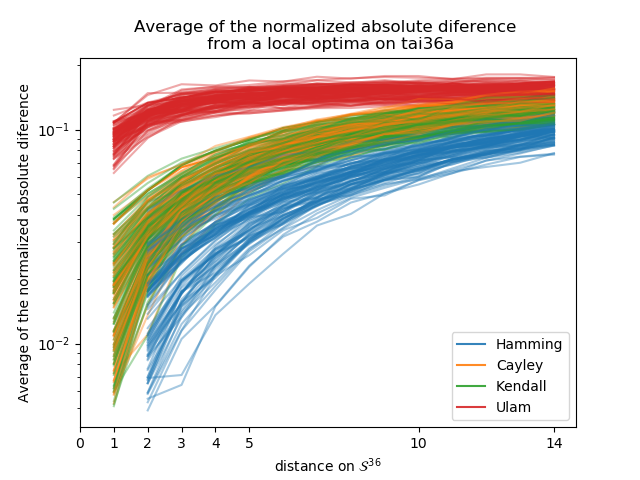} 
			\caption{tai36a}
			\label{fig:bSmoothness}
		\end{subfigure}
		\caption{
			The results of the objective function value transition experiment for two of the instances studied. 
			The instance \textit{bur26a} (a) has special properties on the distance matrix $D$, which we believe makes the Kendall distance have a smoother objective value transition. 
			The instance \textit{tai36a} (b), on the other hand, does not have these properties, and thus, the Hamming distance produces a smoother objective value transition in this case.
		} \label{fig:smoothnessExperiment}
	\end{figure}

	The experiment above demonstrated that, taking the definition of the QAP into account, Hamming is the best option.
	However, considering specific instances of the problem, for many different reasons, the previous conclusion might not hold.	
	For instance, some components could be identical, producing no change; or the change of some components could be compensated by others.
	For that reason, in a new experiment, we will analyze the objective function transition for each of the metrics on specific instances of the problem.
	To that end, starting from a random permutation, we run a local search algorithm\footnote{We used the best-first local search procedure, based on the exchange neighborhood.} to find a local optimum.
	Then, for each of the four distance-metrics, (Hamming, Cayley, Kendall's-$\tau$ and Ulam), the average normalized difference in the objective value with respect to the local optimum is computed for $\forall k \in [14]$.
	Specifically, defining $\sigma_0$ as the local optimum, for each of the metrics, we approximate the difference $\psi_k^{-1} \sum_{\sigma \in \mathcal{S}^n \ | \ d(\sigma_0,\sigma) = k} \frac{\abs(f(\sigma_0) - f(\sigma))}{f(\sigma_0)}$ with the Monte Carlo sampling method using 50 repetitions, where $\psi_k$ is the number of permutations at distance $k$.
	In addition to the average, the variance is also computed.
	The results are deployed\footnote{For this experiment, 6 instances from the QAPLIB are considered.
	Figure~\ref{fig:smoothnessExperiment} shows the results obtained for two of them.
The full results of the experimentation as well as the source code of the proposed approach are available for the interested reader at \url{https://github.com/EtorArza/SupplementaryKMMHamming}} in Figure \ref{fig:smoothnessExperiment}.

	We observed that Hamming shows a smoother objective value transition than Cayley and Ulam, and, most of the times, than Kendall's-$\tau$ (in 4 out of 6 instances) also.
	Considering the results of these two experiments, shown in Figure~\ref{fig:dist_params} and Figure~\ref{fig:smoothnessExperiment}, it seems that the Hamming distance is the best choice among the studied metrics for the QAP.

%

	\section{Distance-Based Probability Models}
	\label{section:mm}
	Probability models for permutations assign a probability value to each of the permutations of n items. 
	For the sake of applicability and computational efficiency, these probability models are defined by a restricted number of parameters. 
	The Mallows Model (MM) \cite{MMcreation} is an exponential probabilistic model defined over $\mathcal{S}^n$. 
	The MM is described by two parameters: the concentration parameter $\theta \in \mathds{R^+}$, and the location parameter, $ \sigma_0 \in \mathcal{S}^n$ \cite{GMM_HAMMING}.
	The location parameter, also known as the central permutation, is the mode of the distribution.
	For the rest of the permutations, their probability decreases exponentially with respect to their distance from the central permutation. 
	The speed of this exponential decay is controlled by $\theta$.
	For instance, when $\theta = 0$, the distribution is equivalent to the uniform distribution over $\mathcal{S}^n$.
	Contrarily, when $\theta \rightarrow \infty$, $p(\sigma_0) = 1$.
	Formally, the probability mass function is given as follows:

	\begin{equation}
	\label{equation:psigma}
	p(\sigma) = p(\sigma | \sigma_0, \theta) = \dfrac{e^{- \theta d(\sigma,\sigma_0)}}{\psi(\theta)} 
	\end{equation}
	
	where $d(\cdot,\cdot)$ is a distance-metric on $\mathcal{S}^n$ and $\psi(\theta)$ stands for the normalization constant.

	\subsection{Factorization and sampling under the Hamming distance}

	In the following, we extend the presentation of the MM for the case of the Hamming distance describing the factorization of the probability distribution induced by the model and the procedure to sample the solutions \cite{GMM_HAMMING}.
	Under this factorization, a simple sampling procedure for the Hamming MM can be defined.
	In addition, this decomposition allows a better understanding of the dynamics of the proposed EDA.
	Defining $\mathcal{K} \equiv d(\sigma_0, \sigma)$ as the Hamming distance from the consensus to $\sigma$, we can think of $\mathcal{K}$ and $\sigma$ as random variables defined in $\{0\}\cup[n]$ and $\mathcal{S}^n$ respectively.
	From this point of view, $\mathcal{K}$ is dependent on $\sigma$, or in other words, given $\sigma$, $\mathcal{K}$ is known.
	Considering this, we can decompose $p(\sigma)$ as:

	\begin{equation}
	\label{equation:psigmadecomp}
	p(\sigma) =  p(\sigma | \mathcal{K}) \ p(\mathcal{K})
	\end{equation}

	where the first term of the factorization, $p(\sigma | \mathcal{K})$, denotes the probability of $\sigma$ given the distance at which it is from the consensus, and the second term, $p(\mathcal{K})$, defines the probability of $k = d(\sigma_0, \sigma)$.
	The conditional probability distribution of the first term of the factorization shown in Equation~\eqref{equation:psigmadecomp}, $p(\sigma | \mathcal{K})$, follows a uniform distribution.
	This is easy to see, since the MM gives the same probability to all permutations that are at the same distance $k$ from the consensus.
	Conveniently, in $\mathcal{S}^n$, the number of permutations at Hamming distance $k$ from a given permutation, $S(n,k)$, can be easily computed.
	This sequence is closely related to the number of derangements of size $k$. 
	A derangement is a permutation, $\sigma$, where every item $\sigma_i$ is different from its corresponding index $i$, hence, $\sigma = (\sigma_1,...,\sigma_k)$ is a derangement of size $k$ iff $\sigma_i \neq i \ , \  \forall i \in [k]$. 
	For example, the permutation, $\tau = (\tau^1, \tau^2, \tau^3) = (3,2,1)$ is not a derangement, because $\tau^2 = 2$, while $\gamma = (\gamma^1,\gamma^2,\gamma^3) =  (3,1,2) $ is a derangement, because $\gamma^i \neq i \ , \ \forall i \in \{1,2,3\}$.

	In order to compute $S(n,k)$, the following formula can be used:
	
	\begin{equation}
	S(n,k) = {n \choose k} D(k)
	\end{equation} 
	
	where $D(k)$ is the number of derangements of size $k$ \cite{irurozki2014r}, which is a known sequence \cite{OEIS:A000166}. 
	Specifically, the number of derangements $D(k)$ can be recursively computed in $\bigO(k)$ as follows:

	\[ D(k) = \begin{cases} 
	1 & k = 0 \\
	0 & k = 1 \\
	(k-1) ( D(k-1) + D(k-2))   & k > 1
	\end{cases}
	\] 
	
	Since we are interested in the first $n+1$ elements of the sequence, we must compute $D(k)$ for $0 \leq k \leq n$, and that requires $\bigO(n)$ time.
	Therefore, it is easy to compute the conditional probability $p(\sigma | \mathcal{K}=k) = S(n,k)^{-1}$ for any $\sigma$ at distance $k$ from the consensus.

	Now, we compute $p(\mathcal{K})$, the second term of the decomposition of $p(\sigma)$ in Equation~\eqref{equation:psigmadecomp}.
	Considering the definition of $p(\mathcal{K}=k) \equiv p(d(\sigma_0,\sigma) = k)$ we obtain:
	
	\begin{equation}
	\label{equation:pd}
	p(\mathcal{K}=k) = \sum_{\sigma | d(\sigma_0,\sigma) = k} p(\sigma)  =  S(n,k) p(\sigma) = S(n,k) \dfrac{e^{- \theta k}}{\psi(\theta)}
	\end{equation}

	The previous equation can be computed in $\bigO(n)$ time for $k \in \{0\}\cup[n]$, allowing a simple two-step sampling procedure for the Hamming MM to be defined \cite{GMM_HAMMING}.
	First, considering the probabilities $p(\mathcal{K})$, randomly choose $k$, the distance at which to sample.
	Secondly, choose a permutation $\sigma$ at distance $k$ from the consensus $\sigma_0$ uniformly at random.
	A detailed explanation of the sampling procedure is shown later, in Section \ref{section:eda}, in Algorithm~\ref{algo:sampleHamming}.

\begin{figure}
		\begin{subfigure}{0.48\textwidth}
			\includegraphics[width=\linewidth]{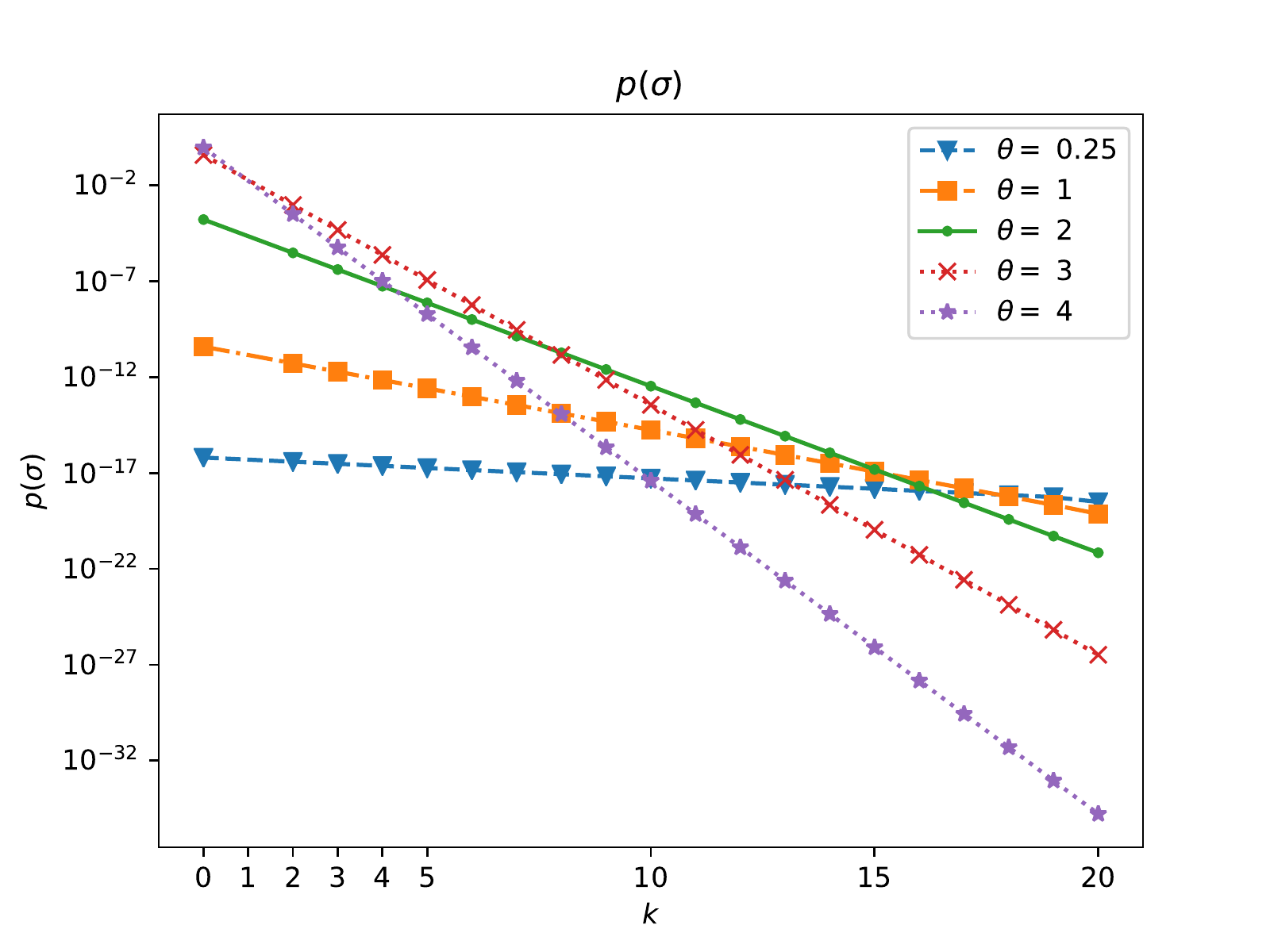} 
			\caption{$p(\sigma)$}
			\label{fig:mono_psigma}
		\end{subfigure}
		\hspace*{\fill} 
		\begin{subfigure}{0.48\textwidth}
			\includegraphics[width=\linewidth]{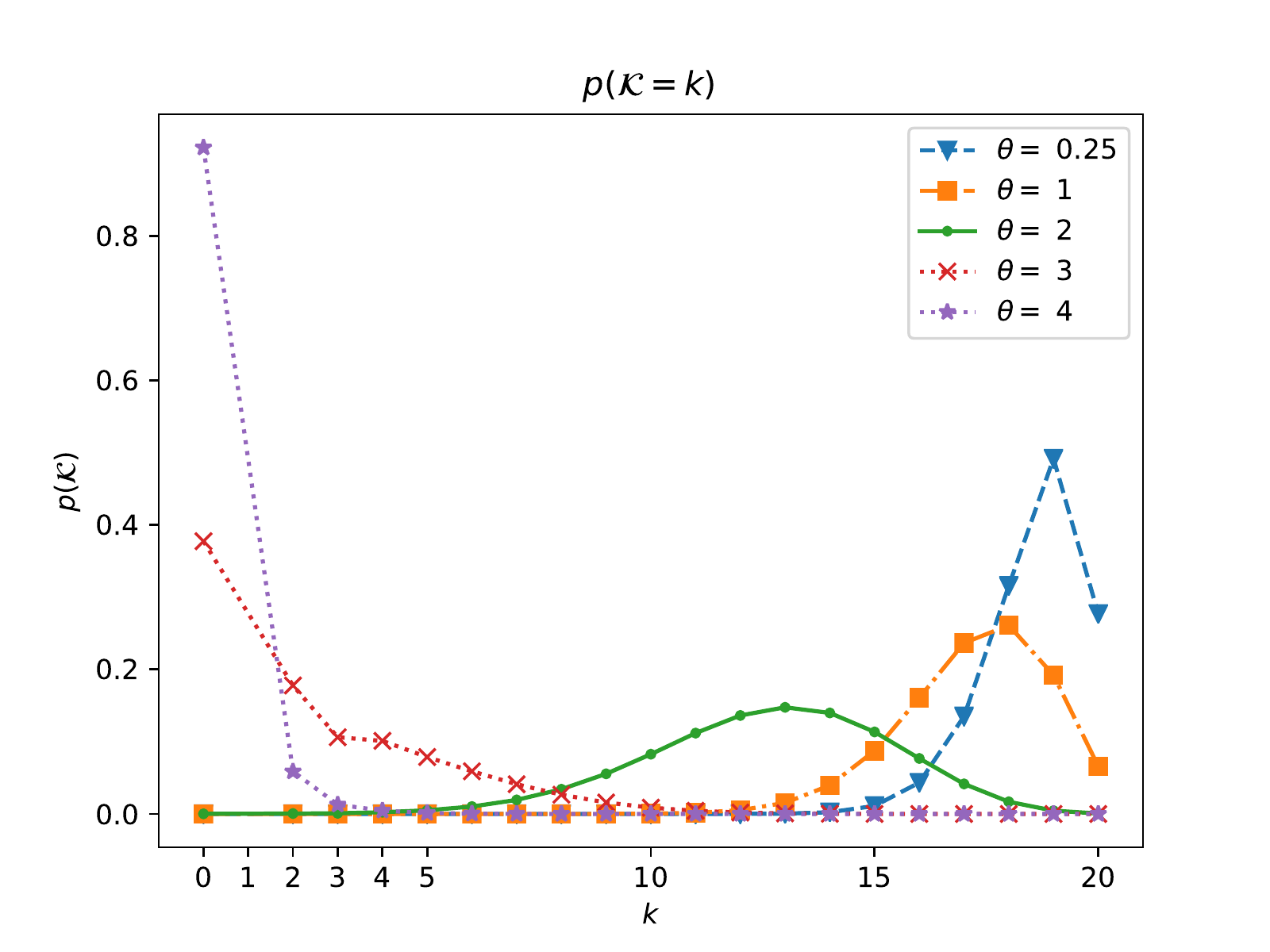} 
			\caption{$p(\mathcal{K})$}
			\label{fig:prob_distance}
		\end{subfigure}
		\caption{
			The representations of $p(\sigma)$ and $p(\mathcal{K})$ on $\mathcal{S}^{20}$ for a Hamming-based MM, considering permutations at Hamming distance $k \in \{0\}\cup[20]$ from the consensus.
			An instance of size $n=20$ and different $\theta$ values are considered.
		} \label{fig:dsigmad}
	\end{figure}

	The concentration parameter $\theta$ controls where the probability of the permutations is concentrated.
	For a high value of $\theta$, the probability mass is concentrated near the consensus.
	Similarly, for a low value of $\theta$, the probability mass is concentrated far away from the consensus.
	This is possible because a low $\theta$ defines an almost uniform distribution on $\mathcal{S}^n$, and the number of permutations at Hamming distance $k$ from the consensus increases exponentially with $k$.
	Figure~\ref{fig:mono_psigma} shows $p(\sigma)$ described in Equation~\eqref{equation:psigma}.
	For high values of $\theta$, $p(\mathcal{K})$ has a higher probability in lower values on $k$, and vice versa.
	In Figure~\ref{fig:prob_distance}, we see that by using different values of $\theta$, the probability mass of $p(\mathcal{K})$ is concentrated at different distances.
	This is related with $\mathbb{E}[\mathcal{K}] = \sum_{k \in \{0\}\cup[n]} k \ p(\mathcal{K}=k)$, the expectation of the distance.
	Given $n$, the instance size, there exists a bijection that maps the expected distance $\mathbb{E}[\mathcal{K}]$ to the corresponding value of the concentration parameter $\theta$.
	When $\mathbb{E}[\mathcal{K}]$ is low, the probability of the solutions near the consensus is high, and, consequently, the concentration parameter of the distribution, $\theta$, is high.
	As we will later see in Section \ref{section:thetaAdjust}, by adjusting $\mathbb{E}[\mathcal{K}]$ (and, consequently, its corresponding $\theta$) a simple exploration-exploitation scheme can be defined.

	\subsection{Extending the Mallows Model}
	The Mallows Model is a unimodal distribution, and as such, it may be too rigid for multimodal problems, limiting the performance of the EDA in certain situations.
	As a more flexible alternative, the Generalized Mallows Model was proposed \cite{GMM} for the Hamming distance \cite{GMM_HAMMING}.
	Based on the decomposition property of some distances, the GMM has a unique central parameter, just as the MM, but it also has several concentration parameters, giving the model a higher flexibility.

	Introduced in \cite{mixturesintroduced}, a multimodal alternative to the MM is the Mixtures of Mallows Model (MMM). 
	In this case, the population is considered to be composed of $m$ differently sized clusters.
	Given the central and concentration parameters for each cluster, $\sigma_i$ and $\theta_i$, the probability mass distribution is expressed as

	$$P(\sigma | \boldsymbol{\sigma},\boldsymbol{\theta}) = \sum_{i=1}^{m} w_i  \dfrac{e^{- \theta_i d(\sigma,\sigma_i)}}{\psi(\theta_i)}$$
	
	where $\sum_{i=1}^{m} w_i = 1 $, $\boldsymbol{w} = \{w_1,...,w_m\}$ and $ w_i > 0$. 
	Usually, $\boldsymbol{\sigma}$, $\boldsymbol{\theta}$ and $\boldsymbol{w}$ are estimated using the Expectation Maximization algorithm \cite{mixturesEMlearning}. 
	An extension of the MMM that considers several concentration parameters per central permutation is the	Mixtures of Generalized Mallows Model (MGMM)~\cite{ceberio2015mixtures}.

	Taking the idea of mixture models to the limit, and by considering each solution in the set as a cluster of equal weight, another even more flexible model can be defined: Kernels of Mallows Model (KMM).
	Given a set of $m$ permutations, the KMM is the averaging of $m$ MMs centered on these permutations. 
	Therefore, we say that KMM is a combination of several MM.
	Contrary to the GMM, the KMM is an MM with the same concentration parameter but different central permutations.
	This model breaks the strong unimodality assumption of the MM and the GMM.
	Given the set of central permutations $\boldsymbol{\sigma} = \{\sigma_1,\sigma_2,...,\sigma_m\}$, the mass probability distribution of KMM can be defined as follows: 
	
	$$P(\sigma | \boldsymbol{\sigma},\theta) = \dfrac{1}{m}  \sum_{i=1}^{m}  \dfrac{e^{- \theta d(\sigma,\sigma_i)}}{\psi(\theta)} $$
	
	where $\psi(\theta)$ is the normalization constant.
%

	%
	%
	%
	%
	%
	%
	%

	\section{Hamming Kernels of Mallows Model EDA}
	\label{section:eda}
	
	In this paper, we approach the QAP with an MM-based EDA.
	Specifically, Kernels of Mallows Model under the Hamming distance are used in the framework of EDAs.
	To control the convergence of the algorithm, a simple yet effective exploration-exploitation scheme is presented, based on $\theta$, the concentration parameter.

	\subsection{Learning and sampling}
	The learning of a model in an EDA usually refers to obtaining the maximum likelihood estimators for the parameters of the selected probabilistic model.
	Since the proposed EDA is based on Kernels of Mallows Model, a non-parametric model, we do not need to estimate the central permutations.
	Instead, the selected set of permutations is used as the set of central permutations $\boldsymbol{\sigma} = D_{t-1}^{sel}$.
	In addition, the concentration parameter $\theta$ is set by a simple exploration-exploitation, as we extensively explain in the next section.



	Once the model is defined, we need to know how to sample solutions from it.
	In this case, the sampling procedure is based on the distances sampling algorithm \cite{irurozki2014r}, as shown in Algorithm~\ref{algo:sampleHamming}.
	It is a three-step procedure.
	First, select a central permutation $\sigma_0$ from the selected set of permutations $\boldsymbol{\sigma}$ uniformly at random (line 1).
	Then, based on the probabilities obtained from Equation~\eqref{equation:pd}, choose a distance $k$ at which to sample (lines 2 and 3).
	Finally, a permutation at Hamming distance $k$ from $\sigma_0$ is chosen uniformly at random (line 4).

	%

	\begin{algorithm}
		
		\DontPrintSemicolon 
		\caption{Sampling algorithm of Hamming KMM}
		\label{algo:sampleHamming}
		\KwIn{\ \\
			\underline{$\boldsymbol{\sigma} = \{\sigma_1,...,\sigma_m\}$}: The set of central permutations.\\
			\noindent \underline{$\theta$}: The concentration parameter.\\ 
		}
		\KwOut{\ \\
			\noindent \underline{$\sigma$}: The sampled permutation.\\ 
		}    
		\SetAlgoLined
		
		$\sigma_0 \gets$ choose uniformly at random from $\boldsymbol{\sigma}$\;
		compute $ \ \left\lbrace p(k) \propto S(n,k) e^{- \theta k} \ | \ k \in [n]\setminus\{1\} \right\rbrace $\;
		$k$ $\gets$ randomly choose $k$ with probability proportional to $p(k)$ (0 is never chosen to avoid sampling the consensus.)\;
		$\sigma \gets$ choose $k$ items from $\sigma_0$ and derange them (shuffle them uniformly at random, but making sure none of these $k$ items remains in its original place)\;
		\textbf{return} $\sigma$\;

	\end{algorithm}

%
%
%
%

	\subsection{Exploration-Exploitation scheme: updating $\theta$}
	\label{section:thetaAdjust}

\begin{figure}
		\begin{subfigure}{0.48\textwidth}
			\includegraphics[width=\linewidth]{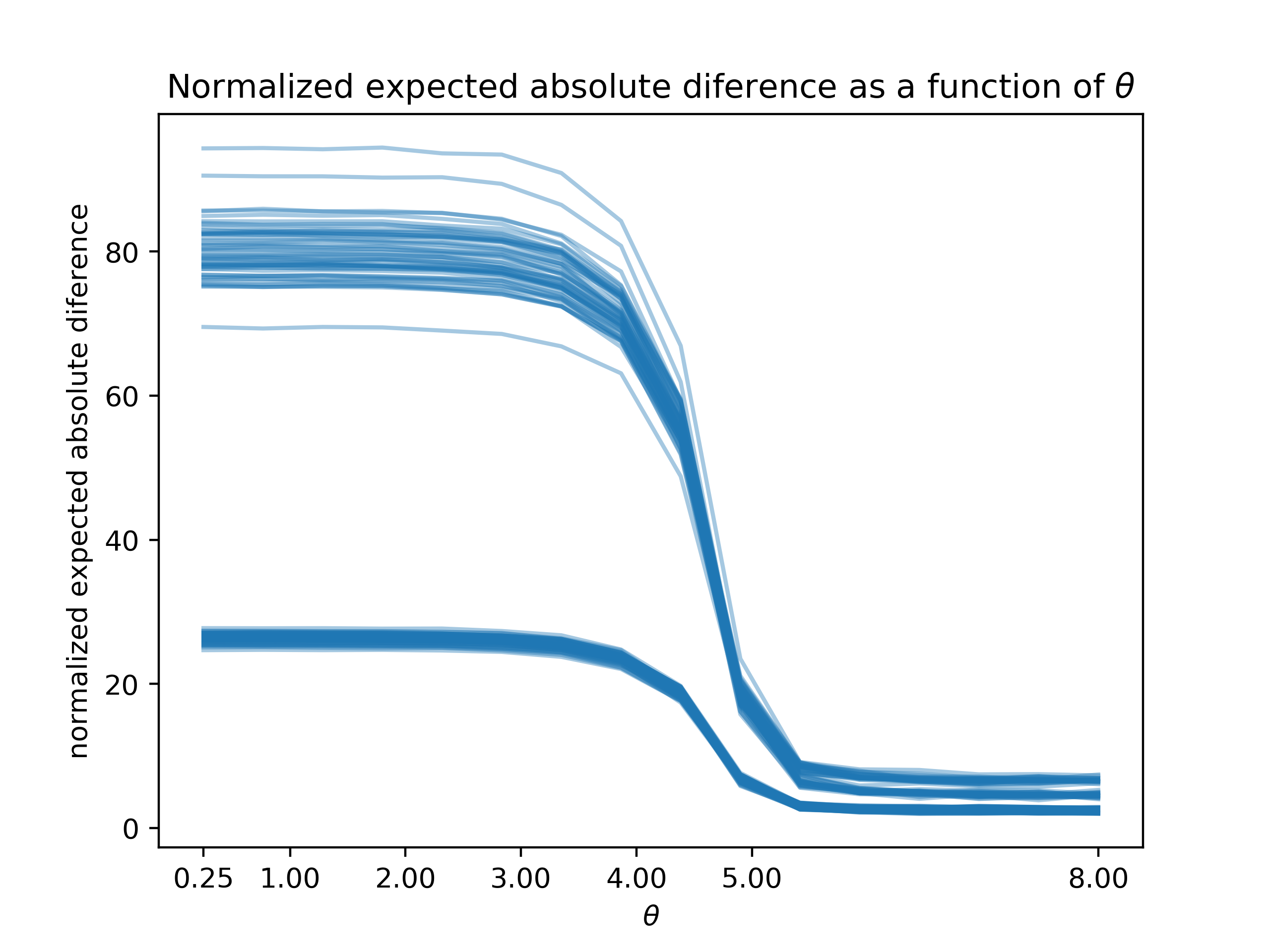}
			\caption{\scriptsize Normalized expected absolute difference between local optima and solutions sampled using the correspondent $\theta$.} \label{fig:a}
		\end{subfigure}
		\hspace*{\fill} 
		\begin{subfigure}{0.48\textwidth}
			\includegraphics[width=\linewidth]{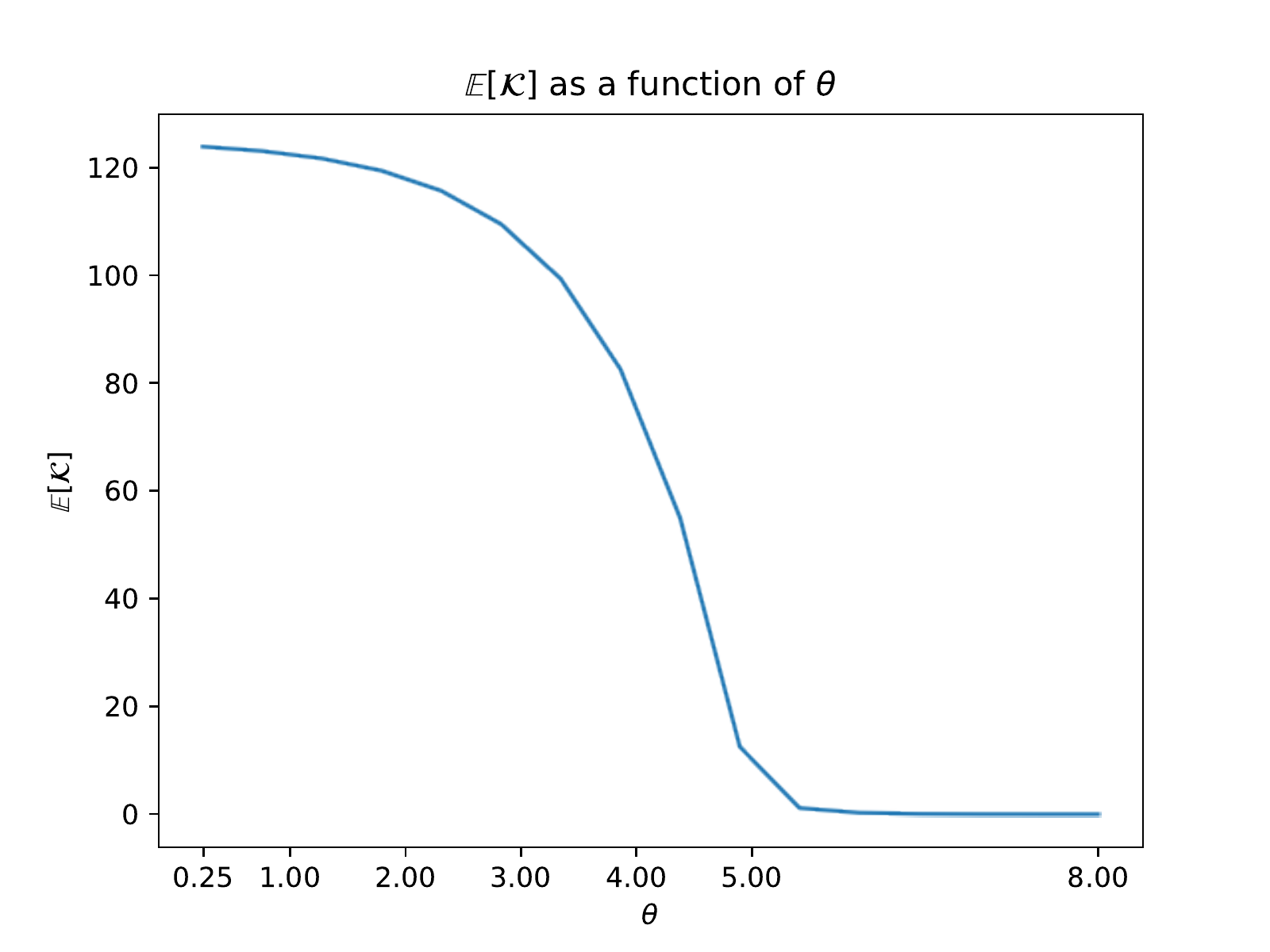}
			\caption{\scriptsize The expectation of $d$, $\mathbb{E}[\mathcal{K}]$ with respect to $\theta$. \vspace{\baselineskip} } \label{fig:b}
		\end{subfigure}
		\caption{
			Figure~\ref{fig:a} shows the normalized expected difference of the objective function value.
			This difference is measured between 100 random local optima and solutions sampled using an MM centered on these local optima and concentration parameter $\theta$.
			Specifically, for each of the considered local optima $\sigma_0$, Figure~\ref{fig:a} shows $\lim\limits_{s \rightarrow \infty} s^{-1} \sum_{i = 1}^{s}  \frac{\abs(f(\sigma_0) - f(\sigma_i))}{f(\sigma_0)}$ where $\sigma_i$ is obtained by sampling from an MM centered on $\sigma_0$ and using the concentration parameter $\theta$ for each $i\in [s]$.
			The instance tai125e01 was used to obtain this figure. 
			Figure~\ref{fig:b} shows the expectation of $d$, $\mathbb{E}[\mathcal{K}]$ as a function of $\theta$.
			Observe how the shape of $\mathbb{E}[\mathcal{K}]$ resembles the normalized expected difference of the objective function value.
		} \label{fig:theta_and_Ed}
	\end{figure}

\begin{figure}
		\centering
		\includegraphics[width=0.48\linewidth]{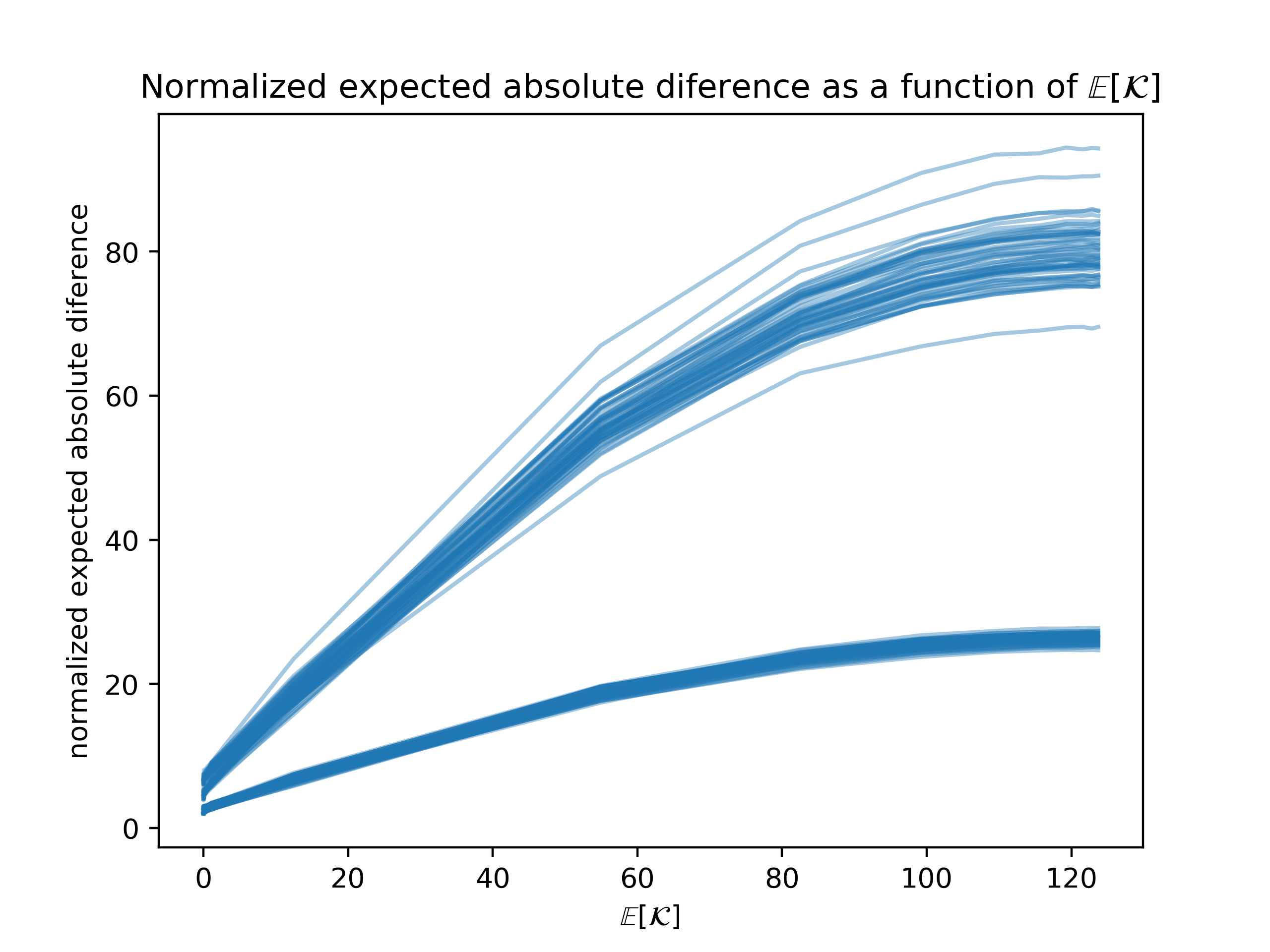}
		\caption{The normalized expected difference of the objective function value as a function of $\mathbb{E}[\mathcal{K}]$.} 
		\label{fig:fitness_as_func_of_expectation}
	\end{figure}

	The convergence of the EDA is controlled by a simple exploration-exploitation scheme.
	The trade-off is balanced by the expectation of the distance, $\mathbb{E}[\mathcal{K}]$, which is transformed into its equivalent $\theta$ at run-time.
	The advantages of using $\mathbb{E}[\mathcal{K}]$ instead of $\theta$ are threefold.
	First, we believe $\mathbb{E}[\mathcal{K}]$ is more intuitive than $\theta$, since its interpretation is much easier.
	In addition, by using $\mathbb{E}[\mathcal{K}]$, it is easier to take into account the instance size $n$ when increasing $\mathbb{E}[\mathcal{K}]$. 
	Finally, $\mathbb{E}[\mathcal{K}]$ is more correlated with the transition of the objective function value than $\theta$.	
	Figure~\ref{fig:theta_and_Ed} shows the evolution of the expected difference in the objective function value and $\mathbb{E}[\mathcal{K}]$ with respect to the concentration parameter $\theta$ for an instance of size $n=125$ (\textit{tai125e01} from the Taixxeyy instances set \cite{taixxinstances}).
	It can be seen that the shape of $\mathbb{E}[\mathcal{K}]$ resembles the normalized expected difference of the objective function.
	In fact, Figure~\ref{fig:fitness_as_func_of_expectation} shows that the relationship between $\mathbb{E}[\mathcal{K}]$ and the normalized objective function difference is almost lineal.
	This means that the transition of the objective value can be more accurately controlled by $\mathbb{E}[\mathcal{K}]$.

	The starting and final values of $\mathbb{E}[\mathcal{K}]_t$\footnote{$\mathbb{E}[\mathcal{K}]_t$ denotes the expectation of the distance $\mathbb{E}[\mathcal{K}]$ at iteration $t$.}, $\mathbb{E}[\mathcal{K}]_{0}$ and $\mathbb{E}[\mathcal{K}]_{tmax}$, respectively, are set before the algorithm is executed.
	In this sense, $\mathbb{E}[\mathcal{K}]_{0}$ is set to a high value.
	A high value of $\mathbb{E}[\mathcal{K}]_t$ favors exploration, because the sampled solutions are expected to be far away from the selected solutions.
	Therefore, the sampled solutions are going to be very different from the selected solutions, forcing them to visit different and unobserved areas of the solution space.	
	At each iteration, the expectation of the distance $\mathbb{E}[\mathcal{K}]$ is decreased ($\mathbb{E}[\mathcal{K}]_{t+1} < \mathbb{E}[\mathcal{K}]_{t}$).
	As the number of iterations increases, the algorithm shifts from an exploration state to an exploitation state.
	In this exploitation stage, the new solutions will be similar (they will be near each other in the Hamming distance sense) to the known solutions.

\begin{figure}
		\centering
		\includegraphics[width=0.48\linewidth]{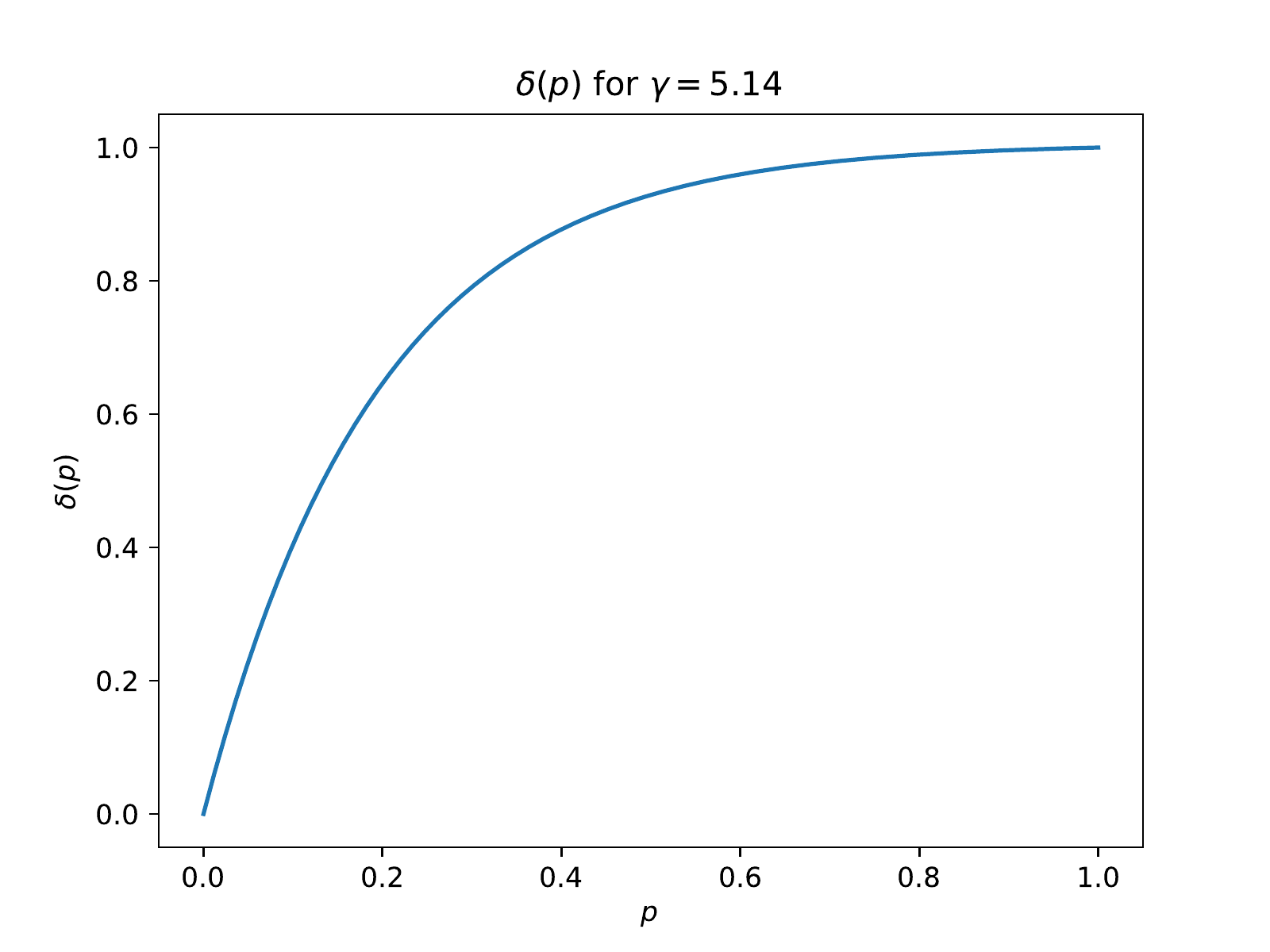}
		\caption{The progress after the exponential function $\delta$ is applied.} 
		\label{fig:ProgressExpo}
	\end{figure}

	An idea to update $\mathbb{E}[\mathcal{K}]_{t}$ would be to decrease it at a constant rate.
	However, we found that decreasing $\mathbb{E}[\mathcal{K}]_{t}$ at an exponential rate produces better results, as we will later discuss in Section~\ref{section:experiments}.
	The stopping criteria for the algorithm is given in terms of the maximum number of iterations, and the number of solutions evaluated in each iteration is $P_s/2$, where $P_s$ denotes the population size of the EDA. 
	Therefore, at each iteration $t$, the progress of the algorithm $p \in (0,1)$ is defined as $p = t / t_{max} $.
	Then, given the intensity parameter $\gamma \in \mathbb{R}^+$, this progress is transformed into an exponential progress with the function $\delta(p) = \frac{\exp(-\gamma p) - 1}{\exp(-\gamma) - 1}$.
	Finally, the expectation at iteration $t$, $\mathbb{E}[\mathcal{K}]_t$, is set to $\mathbb{E}[\mathcal{K}]_t = \mathbb{E}[\mathcal{K}]_{tmax} + \delta(p) (\mathbb{E}[\mathcal{K}]_0  - \mathbb{E}[\mathcal{K}]_{tmax})$.
	Figure~\ref{fig:ProgressExpo} shows $\delta(p)$ for the estimated optimal value of the parameter $\gamma = 5.14$.

	\subsection{Computational complexity and scalability}

	If $n$ is the instance size and $m$ the considered population size, the time complexity of the sampling stage is $\bigO(mn)$ \cite{irurozki2014r}. 
	The total cost of the algorithm, without considering the objective function evaluations, is $\bigO(mn) + \bigO(m\log(m))$. 
	Considering a constant population size $m$, the cost of Hamming KMM EDA is dominated by the cost of the evaluations, which is $\bigO(mn^2)$. 
	The memory complexity of Hamming KMM EDA is $\bigO(mn)$.

\begin{figure}[t]
	\centering
	\includegraphics[width=0.75\linewidth]{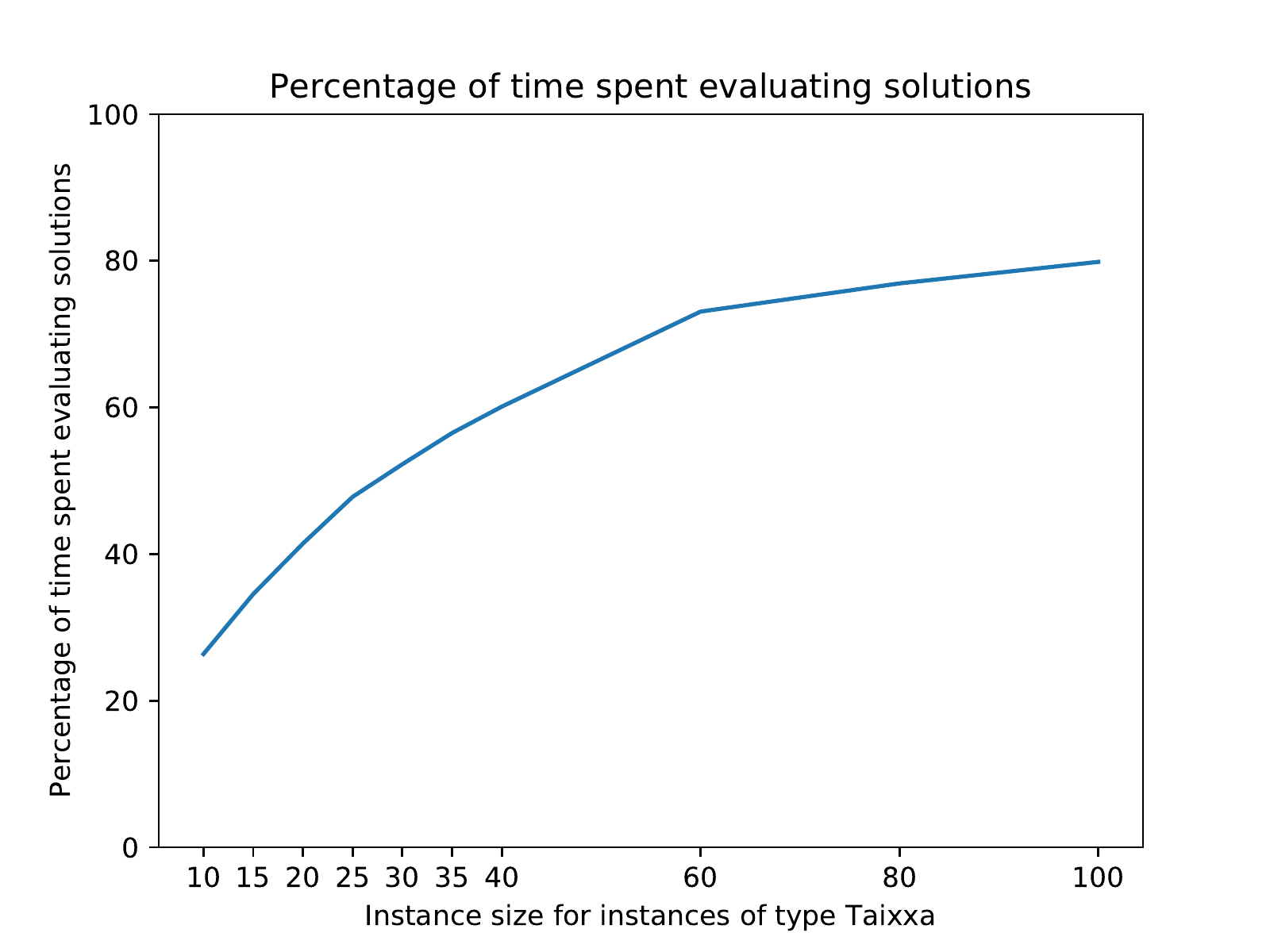}
	\caption{Percentage of time spent on evaluating solutions when optimizing \textit{Taixxa} instances of size $n$, with $1000n^2$ as stopping criterion.} 
	\label{fig:percentage_of_time}
\end{figure}

	Even though the theoretical computational complexity of Hamming KMM EDA is reasonable (since it is dominated by the cost of the evaluations, $\bigO(m \cdot n^2)$), a detailed empirical analysis is recommended to study the scalability of the algorithm.
	To this end, we conducted two experiments on the runtime of the proposed algorithm.
	In the first experiment, we computed the percentage of time spent evaluating the objective function when using Hamming KMM to optimize \textit{Taixxa} instances of size $n$, with $n$ ranging from 10 to 100.
	The stopping criterion was set to $1000 \cdot n^2$ evaluations, and the rest of the parameters remain equal to those considered in the experimental section.
	Results are depicted in Figure~\ref{fig:percentage_of_time}. 
	As can be observed, as the instance size increases, the percentage of time spent evaluating the objective function also increases, up to $80\%$ in instances of size $100$.
	This means that the runtime of Hamming KMM is indeed dominated by the evaluation of the objective function, which is a positive feature, since it means that the runtime overhead of the EDA is small in comparison to the cost of evaluating the solutions.

\begin{figure}[t]
	\centering
	\includegraphics[width=0.75\linewidth]{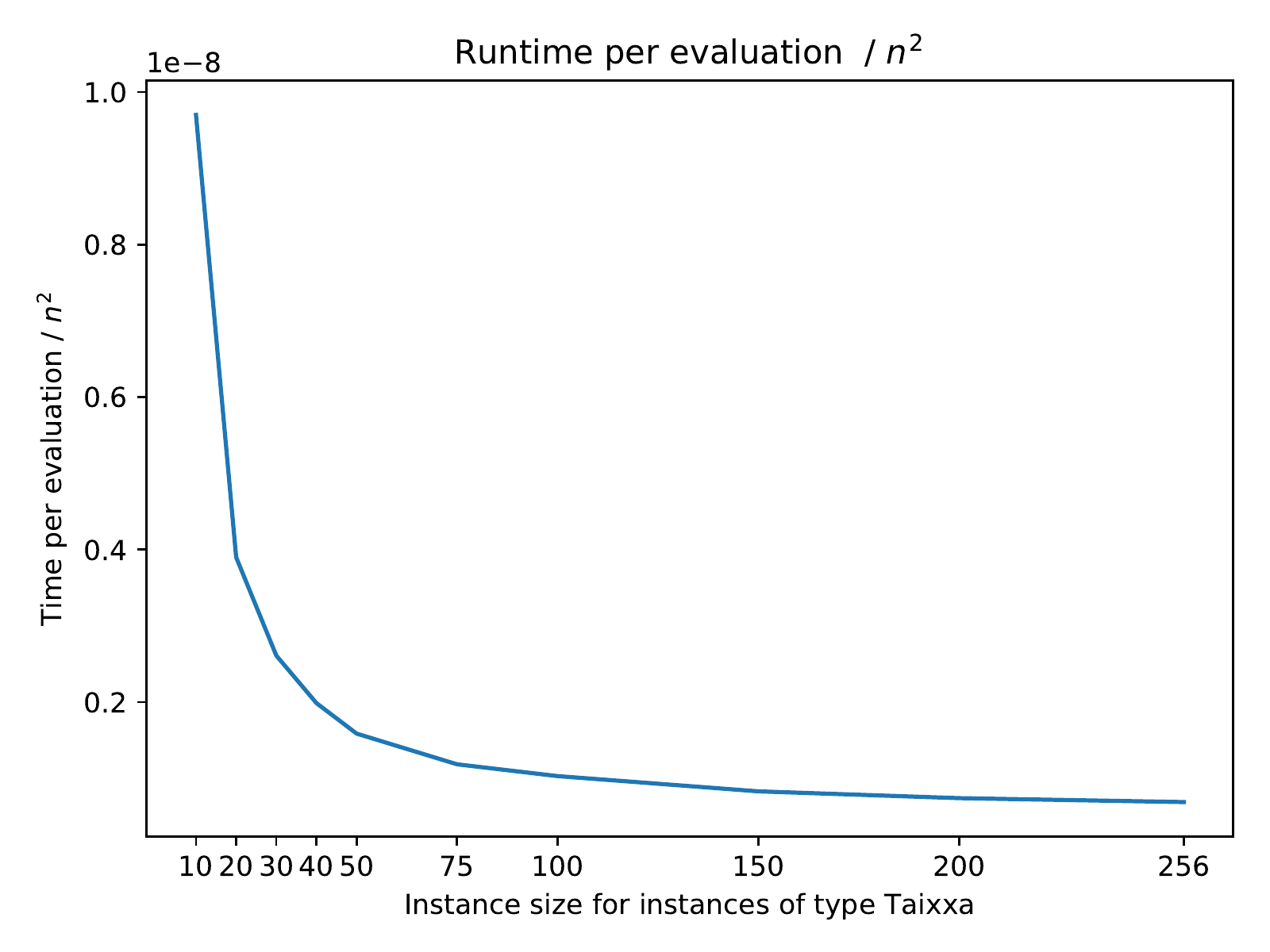}
	\caption{Runtime per evaluation divided by $n^2$ of Hamming KMM when optimizing \textit{Taixxc} instances of size $n$, with $10^6$ evaluations as stopping criterion.} 
	\label{fig:time_per_ev_divided}
\end{figure}

	In order to further prove this point, an additional experiment was conducted.
	Specifically, in this second experiment, we empirically show that the cost of the Hamming KMM EDA algorithm is at most $\bigO(n^2)$ for a fixed population size of $m = 972$, where $n$ is the problem instance size.
	With this aim in mind, we run Hamming KMM in instances obtained by cutting \textit{Tai256c}, the largest instance available in the QAPLIB~\cite{burkard1997qaplib}.
	In this experiment, the stopping criterion was set to $10^6$ evaluations, and the rest of the parameters remain equal to those considered in the experimental section.
	In Figure~\ref{fig:time_per_ev_divided}, the proportional time spent on each function evaluation divided by $n^2$ is shown.
	Specifically, $\frac{\textit{runtime}}{E \cdot n^2}$ is shown, where $E$ is the number of evaluations used as stopping criterion ($E = 10^6$), and \textit{runtime} is the time it takes to optimize an instance of size $n$.
	Results point out that $\frac{\textit{runtime}}{E \cdot n^2}$ decreases as $n$ increases.
	This means that for a fixed stopping criterion in terms of maximum evaluations and a fixed population size, the actual time complexity of Hamming KMM is at most $\bigO(n^2)$.
	In the following, we show how this cost can be reduced even further.

	Even though the cost of evaluating a candidate solution is $\bigO(n^2)$, given two different permutations $\sigma_a, \sigma_b \in \mathcal{S}^n$, if $\sigma_b \in \mathcal{N}(\sigma_a)$ and the objective function value of $\sigma_a$ is known, then the objective function value of $\sigma_b$ can be updated in $\bigO(n)$ time \cite{FastSwap}.
	The proposition below defines the objective function relationship that two solutions at Hamming distance two have.

	\begin{prop}

	Suppose $\exists i_1,i_2 \in [n] | \sigma_a(i) = \sigma_b(i) \ \ \forall i \in [n] \setminus \{i_1,i_2\}$. Then, \\
$$f(\sigma_b)~=~f(\sigma_a)+ 
\sum\limits_{k \in \{i_1,i_2\}} \sum\limits_{i=0}^{n-1} \left( 
D_{i,k} H_{\sigma_b(i), \sigma_b(k)} 
+  D_{k,i} H_{\sigma_b(k), \sigma_b(i)} \right)- $$ $$
\sum\limits_{k \in \{i_1,i_2\}} \sum\limits_{i=0}^{n-1} \left( 
D_{i,k} H_{\sigma_a(i), \sigma_a(k)} 
+  D_{k,i} H_{\sigma_a(k), \sigma_a(i)} \right) +
\sum\limits_{k \in \{i_1,i_2\}}  D_{k,k} H_{\sigma_a(k), \sigma_a(k)} - D_{k,k} H_{\sigma_b(k), \sigma_b(k)}$$  \vspace{0.1cm}

	\end{prop}

	This process can be repeated over and over again to compute the objective function value of permutations at Hamming distance two or more, and if the permutations are close enough, it is more efficient than directly computing $f(\sigma_b)$.
	It is worth noting that the proposed approach is based on MM kernels and we used the Distances Sampling Algorithm as the sampling procedure \cite{irurozki2014r}. 
	Hence, we sample at a given distance of a known permutation. 
	Therefore, this efficient method to compute the objective function yields a considerable speedup in the EDA, especially in the last iterations of the EDA, where the expected distance from the central permutation to the sampled permutation $\mathbb{E}[\mathcal{K}]$ is small.

	\section{Experimental study}
	\label{section:experiments}
	In order to prove the validity of the proposed method, in this section, we present an exhaustive analysis of the performance of the algorithm.

	\subsection{General remarks}

	Some popular instances of the QAP have been employed to evaluate the performance of the proposed approach (Hamming KMM EDA). 
	In particular, 30 instances from the QAPLIB~\cite{burkard1997qaplib} were used.

	Before running the experiments, there are a number of parameters that need to be set in the proposed approach.
	First, we have the starting $\mathbb{E}[\mathcal{K}]_0$ and final $\mathbb{E}[\mathcal{K}]_{max}$ values of $\mathbb{E}[\mathcal{K}]$.
	$\mathbb{E}[\mathcal{K}]_0$ is set to $n/2$, where $n$ is the instance size, and $\mathbb{E}[\mathcal{K}]_{max}$ is set to $0.25$.
	Setting $\mathbb{E}[\mathcal{K}]_0$ to $n/2$ produces a distribution in which, on average, the sampled permutations have half the items in the same position as the reference permutation $\sigma_0$.
	The chosen $\mathbb{E}[\mathcal{K}]_{max}$ value produces a similar distribution that $\mathbb{E}[\mathcal{K}]_{max} \rightarrow 0$ would, but without numerical errors, it is thus the most exploitative state possible.
	The other two parameters are $P_s$ and $\gamma$.
	The parameter $P_s$ is the population size of the EDA, and $\gamma$ measures the speed at which $\mathbb{E}[\mathcal{K}]$ is decreased.
	These two parameters are set using Bayesian optimization \cite{scikit-learn} with the instance \textit{tai31a} (which is not among the benchmark instances considered).
	The optimal values found for these parameters are 972 and 5.14 respectively.
	These parameters are used in all the executions of Hamming KMM EDA.

	All the algorithms considered in the experimentation are tested in the set of 30 instances.
	The stopping criterion is the same for all the considered algorithms and instances: $1000n^2$ evaluations.
	The experimentation was conducted on a single machine with an octa-core AMD Ryzen 7 1800X Processor, with 8Gb of RAM.
	Hamming KMM was implemented in C++.
	The rest of the algorithms were implemented in either Java or C++.
	In any case, since the number of evaluations is used as the stopping criteria, it is not affected by the hardware nor the programming language, and therefore, is easy to reproduce.
	As a reference, it takes Hamming KMM about 107 seconds to perform $1000n^2$ evaluations in the largest of the studied instances (tai100a) and 0.1 seconds in the smallest one (tai10a).

	For each benchmark instance, the results are recorded as the Average Relative Deviation Percentage, ARDP $=|\frac{f_{best} - f_{av}}{f_{best}}|$, where $f_{best}$ is the best known value and $f_{av}$ is the average of the best objective values obtained in each repetition.
	For further statistical analysis, Bayesian Performance Analysis~\cite{BayesianPLPerformance,calvo2016scmamp} (BPA) is carried out to study the uncertainty of the results of each experiment\footnote{In the file  \href{https://github.com/EtorArza/SupplementaryKMMHamming/blob/master/comparison_between_BPA_and_NHST.pdf}{"comparison\_between\_BPA\_and\_NHST.pdf"}, available in the \href{https://github.com/EtorArza/SupplementaryKMMHamming}{GitHub repository}, we justify the use of Bayesian performance analysis instead of other more traditional hypothesis tests.}
	Specifically, Placket-Luce is used as the probability model, defined in $\mathcal{S}^n$, in this case, corresponding to the rankings of the algorithms.
	BPA considers probability distributions over probability distributions.
	In our case, assuming a uniform prior distribution, the posterior distribution of the probability of each algorithm being the best performing one (winning) is computed.
	The goal of this analysis is to determine which algorithm performs the best for the set of considered benchmark instances.
	The inference with the BPA approach is fairly simple~\cite{BayesianPLPerformance}.
	First the scores of the algorithms are transformed from ARDP to their corresponding rankings on each of the test instances.
	Then, a sample is produced from the posterior probability distribution of the weights of the Plackett-Luce model (a probability model for rankings).
	And finally, based on these sampled weights, the credible interval of $90\%$ is computed for each algorithm.
	This interval means that there is a $90\%$ chance that the actual probability of the algorithm being the highest ranking algorithm (being the winner) lies within the interval.


	\color{black}

\subsection{Experiment 1: Kernels and the exponential $\mathbb{E}[\mathcal{K}]$}
\begin{figure}
	\centering
	\includegraphics[width=0.38\linewidth]{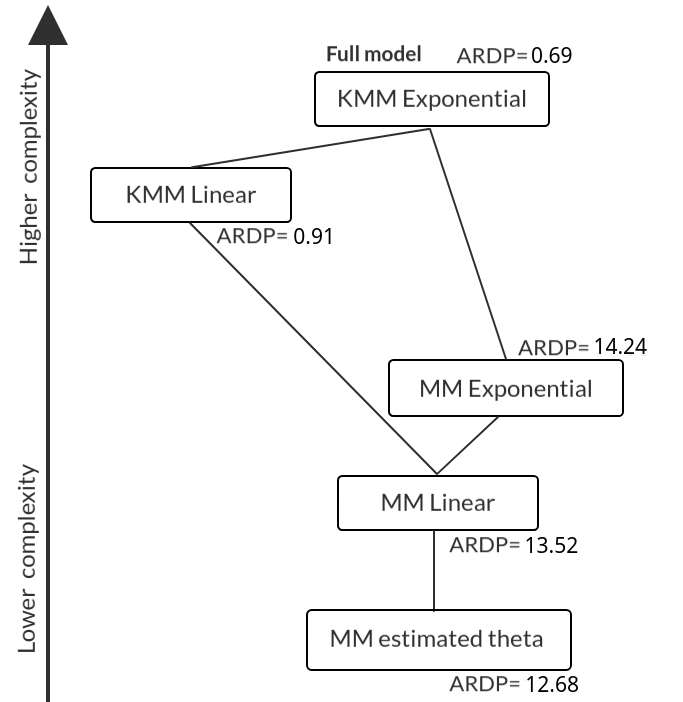}
	\caption{A diagram of the simplified models considered in this paper. 
		The vertical axis is proportional the complexity level of the model in terms of the number of free parameters.
		The average ARDP obtained in the studied instances is shown for each model.} 
	\label{fig:ComplexityDiagram}
\end{figure}

For the sake of measuring the contribution of each of the two main parts that extend a Hamming Mallows Model EDA, (i) the use of kernels and (ii) the use of an exponential increase of $\mathbb{E}[\mathcal{K}]$, we compare the performance of the full model with the simplified variants.
The simplified models considered are: KMM with linear increase of $\mathbb{E}[\mathcal{K}]$, MM with exponential increase of $\mathbb{E}[\mathcal{K}]$, both with one missing part; MM with linear increase of $\mathbb{E}[\mathcal{K}]$, missing both parts; and finally, a simple Hamming MM in which the concentration parameter $\theta$ is estimated at each iteration. 
Figure~\ref{fig:ComplexityDiagram} shows all the studied simplified models, ordered by their complexity in terms of the number of free parameters.
The average ARDP obtained in all the instances for each of the models is also shown in this figure.

The same parameters are considered for all the EDAs, thus, the parameters estimated with Bayesian optimization for the full model are used.
The ARDP values are recorded in Table \ref{table:Complexity}.

The full model outperforms the rest of the models in $60\%$ of the instances, while the second best model, (KMM with linear increase), only outperforms the rest of the models in $23.3\%$ of the studied instances.
It is worth noting that using the kernels part is much more important than the exponential increase of $\mathbb{E}[\mathcal{K}]$, since, as seen in Figure~\ref{fig:ComplexityDiagram}, both kernel models have an average ARDP lower than $1\%$, while the rest of the models have an ARDP over $10\%$. 
Additionally, the exponential increase is detrimental for the MM, and it is only a positive addition when we consider it alongside the kernels parts.
This experiment indicates that both the kernels and the exponential increase of $\mathbb{E}[\mathcal{K}]$ are key parts of the proposed model.

\begin{figure}
	\centering
    \textbf{Probability of winning for the simplified models}\par\medskip
	\includegraphics[width=0.5\linewidth]{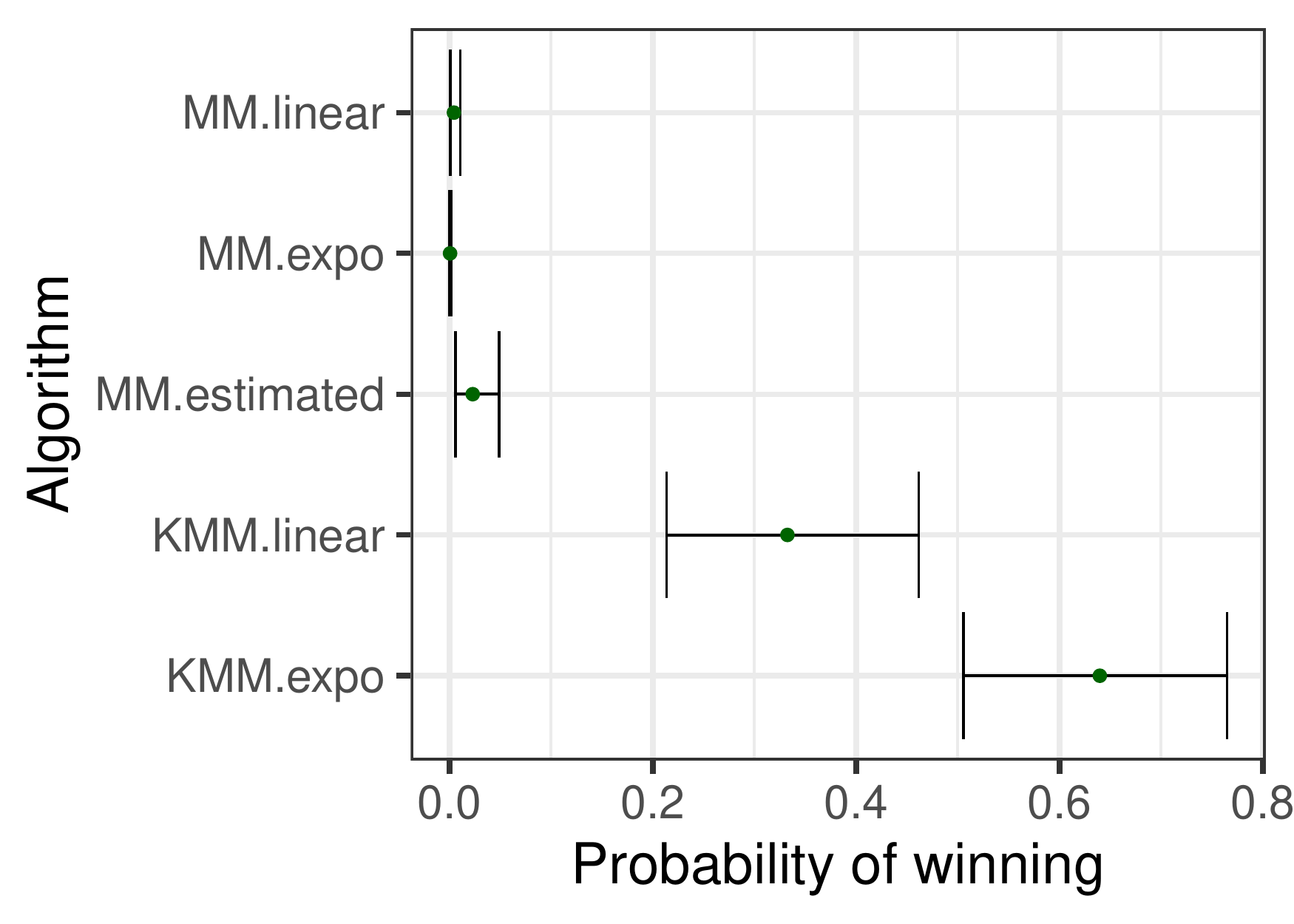}
	\caption{Credible intervals of 90\% and expected value of the estimated posterior probability of each algorithm being the winner among those tested.} 
	\label{fig:ablations_prob_of_wining_09}
\end{figure}

Figure~\ref{fig:ablations_prob_of_wining_09} shows the credible interval of $90\%$ calculated from the samples obtained from the posterior distribution of the probability of each of the algorithms being the best.
As can be observed, both kernel-based algorithms (with exponential increase and linear increase) have a higher expected probability of being the best than the rest, around 0.6 and 0.4 respectively.

\setlength{\tabcolsep}{4pt}


\begin{table}
	\begin{center}
	\begin{scriptsize}
	\caption{The results of the simplified models when compared with the full model. The best performing algorithm is highlighted in bold.}
	\label{table:Complexity}
	\begin{tabular}{lrrrrr} 
Instance & KMM Exponential & KMM Linear & MM Exponential & MM Linear & MM estimated $\theta$\\
\hline
bur26a & 0.105 & \textbf{0.100} & 1.600 & 1.469 & 1.420 \\
bur26b & 0.182 & \textbf{0.165} & 1.596 & 1.534 & 1.335 \\
bur26c & \textbf{0.007} & 0.010 & 2.202 & 1.970 & 1.717 \\
bur26d & \textbf{0.007} & \textbf{0.007} & 2.342 & 2.147 & 1.930 \\
nug17 & 0.179 & \textbf{0.110} & 9.919 & 9.371 & 8.655 \\
nug18 & \textbf{0.326} & 0.409 & 10.415 & 10.332 & 9.798 \\
nug20 & \textbf{0.125} & 0.175 & 11.008 & 10.911 & 10.381 \\
nug21 & 0.271 & \textbf{0.197} & 14.475 & 13.515 & 12.806 \\
tai10a & \textbf{0.000} & \textbf{0.000} & 3.484 & 3.207 & 2.338 \\
tai10b & \textbf{0.000} & \textbf{0.000} & 2.814 & 2.653 & 1.524 \\
tai12a & 0.140 & \textbf{0.000} & 9.624 & 8.806 & 8.280 \\
tai12b & \textbf{0.000} & \textbf{0.000} & 4.924 & 4.974 & 4.635 \\
tai15a & \textbf{0.179} & 0.190 & 8.252 & 8.194 & 7.536 \\
tai15b & \textbf{0.007} & \textbf{0.007} & 0.910 & 1.020 & 0.822 \\
tai20a & \textbf{0.843} & 0.947 & 12.830 & 12.379 & 11.863 \\
tai20b & \textbf{0.068} & 0.123 & 9.349 & 9.089 & 9.165 \\
tai25a & \textbf{1.265} & 1.732 & 13.586 & 12.941 & 12.454 \\
tai25b & \textbf{0.025} & 0.041 & 30.511 & 27.680 & 21.286 \\
tai30a & \textbf{1.435} & 1.891 & 12.547 & 12.258 & 11.957 \\
tai30b & 0.189 & \textbf{0.075} & 34.105 & 29.592 & 23.616 \\
tai35a & \textbf{1.485} & 2.429 & 13.852 & 13.121 & 12.800 \\
tai35b & \textbf{0.476} & 0.526 & 30.253 & 27.398 & 24.357 \\
tai40a & \textbf{1.762} & 2.622 & 13.829 & 13.451 & 13.157 \\
tai40b & 1.068 & \textbf{0.299} & 37.258 & 34.402 & 33.339 \\
tai60a & \textbf{2.237} & 3.400 & 13.757 & 13.524 & 13.180 \\
tai60b & \textbf{0.493} & 0.647 & 33.766 & 32.900 & 33.237 \\
tai80a & \textbf{2.172} & 3.658 & 12.547 & 12.361 & 12.005 \\
tai80b & \textbf{2.235} & 2.707 & 33.004 & 32.617 & 32.004 \\
tai100a & \textbf{2.190} & 3.538 & 11.771 & 11.619 & 11.485 \\
tai100b & \textbf{1.142} & 1.404 & 30.624 & 30.156 & 31.466 \\

		\end{tabular}
			\end{scriptsize}
	\end{center}
\end{table}

\subsection{Experiment 2: Comparing specific EDAs for permutation problems}

In this experiment, we compare Hamming KMM EDA to other specific EDAs for permutation problems\footnote{Specific EDAs are those that estimate a probability distribution explicitly on $\mathcal{S}_n$.} considered in the literature.
We compare the performance of the proposed approach with respect to other MM-based approaches.
For example, the MM has already been applied to permutation problems under the Cayley, \mbox{Kendall's-$\tau$} and Ulam distances \cite{review_on_distances}.
Cayley and Kendall-based KMMs \cite{ceberio2015kernels} and GMMs have also been studied.
Additionally, \textit{mixtures of GMMs} (MGMMs) have also been applied to permutation-based problems under the Cayley and Kendall distances \cite{ceberio2015mixtures}. 
Specifically, in this last article, the MGMM with two clusters was found to outperform the other MGMM approaches, and therefore, we will only consider MGMMs with two clusters.
In addition to MM EDAs, we compare the performance of the proposed algorithm to a Plackett-Luce EDA \cite{ceberio_plackett-luce_2013}.

The ARDP value for all the instances is recorded in Table \ref{table:otherMM}.
We kept the parameters proposed by each author in the paper that the algorithm is proposed.

\setlength{\tabcolsep}{4pt}
\newcommand*{\myalign}[2]{\multicolumn{1}{#1}{\hspace{0.2cm}#2}}

\begin{table}
	\begin{center}
		\begin{scriptsize}
			\caption{The average ARPD results of Hamming KMM EDA and other EDA approaches specific to $S^n$. The best performing algorithm is highlighted in bold.}
			\label{table:otherMM}
			\begin{tabular}{lcrrrrrcrrrrr} 	
				& \tiny  Hamming & \tiny  Ulam & \multicolumn{4}{c}{\tiny Kendall} && \multicolumn{4}{c}{\tiny Cayley}  & \tiny Plackett-	\\
				\cline{4-7} \cline{9-12}
				\rule{0pt}{3ex}    
				Instance & \myalign{c}{\tiny KMM} & \myalign{c}{\tiny MM} & \myalign{c}{\tiny MM} & \myalign{c}{\tiny KMM} & \myalign{c}{\tiny MGMM} & \myalign{c}{\tiny GMM} & \myalign{c}{ } & \myalign{c}{\tiny MM} & \myalign{c}{\tiny KMM} & \myalign{c}{\tiny MGMM} & \myalign{c}{\tiny GMM} & \myalign{c}{\tiny Luce}\\
				& \tiny  &\myalign{c}{ \cite{review_on_distances}} & \myalign{c}{ \cite{review_on_distances}} & \myalign{c}{\cite{ceberio2015kernels}}& \myalign{c}{\cite{ceberio2015mixtures}} & \myalign{c}{\cite{ceberio_distance-based_2014}} && \myalign{c}{\cite{review_on_distances}} & \myalign{c}{\cite{ceberio2015kernels}} & \myalign{c}{\cite{ceberio2015mixtures}} & \myalign{c}{\cite{ceberio_distance-based_2014}} &  \myalign{c}{\cite{ceberio_plackett-luce_2013}}\\
				\hline
				\\
bur26a & \textbf{0.105} & 3.716 & 1.937 & 2.453 & 1.955 & 1.746 && 0.359 & 0.211 & 0.690 & 0.365 & 1.730 \\
bur26b & \textbf{0.182} & 4.052 & 2.008 & 2.353 & 1.998 & 1.461 && 0.482 & 0.305 & 0.603 & 0.465 & 1.680 \\
bur26c & \textbf{0.007} & 4.386 & 2.121 & 2.655 & 2.054 & 1.827 && 0.369 & 0.172 & 0.634 & 0.307 & 1.843 \\
bur26d & \textbf{0.007} & 4.756 & 2.352 & 2.872 & 2.233 & 1.882 && 0.370 & 0.168 & 0.695 & 0.355 & 1.890 \\
nug17 & \textbf{0.179} & 9.434 & 9.174 & 12.535 & 9.284 & 7.771 && 4.706 & 2.154 & 6.680 & 3.147 & 7.188 \\
nug18 & \textbf{0.326} & 17.601 & 8.881 & 12.746 & 10.306 & 8.316 && 4.689 & 2.948 & 6.161 & 3.684 & 7.850 \\
nug20 & \textbf{0.125} & 11.391 & 9.603 & 11.961 & 8.541 & 7.759 && 5.195 & 1.844 & 2.977 & 3.214 & 10.895 \\
nug21 & \textbf{0.271} & 13.934 & 12.252 & 12.613 & 12.137 & 10.738 && 5.722 & 2.695 & 7.859 & 3.380 & 13.659 \\
tai10a & \textbf{0.000} & 15.514 & 6.464 & 11.306 & 8.048 & 8.395 && 2.354 & 1.038 & 2.244 & 2.348 & 2.278 \\
tai10b & \textbf{0.000} & 25.309 & 7.025 & 15.845 & 7.506 & 6.327 && 1.032 & 1.440 & 1.475 & 2.324 & 2.835 \\
tai12a & \textbf{0.140} & 9.521 & 11.905 & 17.115 & 12.359 & 11.496 && 7.226 & 5.856 & 7.415 & 6.415 & 6.928 \\
tai12b & \textbf{0.000} & 6.043 & 11.716 & 21.591 & 13.287 & 11.565 && 5.046 & 3.551 & 8.264 & 6.414 & 6.757 \\
tai15a & \textbf{0.179} & 7.919 & 9.225 & 11.174 & 9.670 & 9.128 && 4.648 & 3.153 & 6.497 & 3.586 & 5.631 \\
tai15b & \textbf{0.007} & 0.975 & 1.279 & 1.536 & 1.224 & 0.997 && 0.528 & 0.369 & 0.749 & 0.411 & 0.743 \\
tai20a & \textbf{0.843} & 12.014 & 12.050 & 12.921 & 11.443 & 10.942 && 6.971 & 2.820 & 5.412 & 5.355 & 11.958 \\
tai20b & \textbf{0.068} & 7.494 & 13.348 & 32.965 & 12.735 & 12.788 && 5.112 & 5.322 & 2.345 & 2.646 & 6.912 \\
tai25a & \textbf{1.265} & 12.355 & 11.856 & 12.439 & 11.765 & 11.159 && 7.572 & 4.735 & 8.397 & 5.537 & 11.962 \\
tai25b & \textbf{0.025} & 16.446 & 24.200 & 56.513 & 30.102 & 22.254 && 6.071 & 5.456 & 10.529 & 4.692 & 20.214 \\
tai30a & \textbf{1.435} & 15.292 & 11.263 & 11.314 & 10.529 & 10.184 && 6.628 & 3.301 & 4.629 & 4.947 & 11.682 \\
tai30b & \textbf{0.189} & 49.972 & 29.100 & 45.465 & 27.863 & 21.636 && 9.282 & 7.843 & 10.025 & 11.077 & 22.984 \\
tai35a & \textbf{1.485} & 12.912 & 11.879 & 11.820 & 11.862 & 11.221 && 7.316 & 4.592 & 7.621 & 4.880 & 12.921 \\
tai35b & \textbf{0.476} & 20.097 & 24.583 & 36.410 & 29.819 & 21.667 && 7.083 & 5.976 & 9.532 & 5.346 & 25.320 \\
tai40a & \textbf{1.762} & 13.257 & 11.651 & 11.546 & 11.599 & 11.004 && 7.162 & 3.670 & 4.904 & 4.862 & 13.272 \\
tai40b & \textbf{1.068} & 28.524 & 30.422 & 44.129 & 33.423 & 25.576 && 10.729 & 8.421 & 6.970 & 8.703 & 33.436 \\
tai60a & \textbf{2.237} & 13.222 & 11.367 & 10.996 & 11.026 & 10.234 && 7.354 & 3.878 & 4.666 & 4.574 & 13.103 \\
tai60b & \textbf{0.493} & 34.853 & 33.144 & 40.119 & 33.344 & 24.317 && 7.112 & 5.491 & 5.897 & 5.238 & 32.017 \\
tai80a & \textbf{2.172} & 12.109 & 9.964 & 9.740 & 9.853 & 9.465 && 6.525 & 3.745 & 4.111 & 4.358 & 12.074 \\
tai80b & \textbf{2.235} & 32.741 & 31.416 & 33.977 & 30.511 & 26.626 && 6.674 & 5.295 & 6.053 & 5.792 & 32.458 \\
tai100a & \textbf{2.190} & 11.496 & 9.469 & 9.062 & 9.282 & 8.711 && 6.237 & 3.460 & 3.617 & 3.913 & 11.469 \\
tai100b & \textbf{1.142} & 31.929 & 25.849 & 31.703 & 29.561 & 22.020 && 5.469 & 4.603 & 4.888 & 4.982 & 31.116 \\

\end{tabular}
		\end{scriptsize}
	\end{center}
\end{table}

\begin{figure}
	\centering
	\textbf{Probability of winning for EDAs specific to $\mathcal{S}^n$}\par\medskip
	\includegraphics[width=0.5\linewidth]{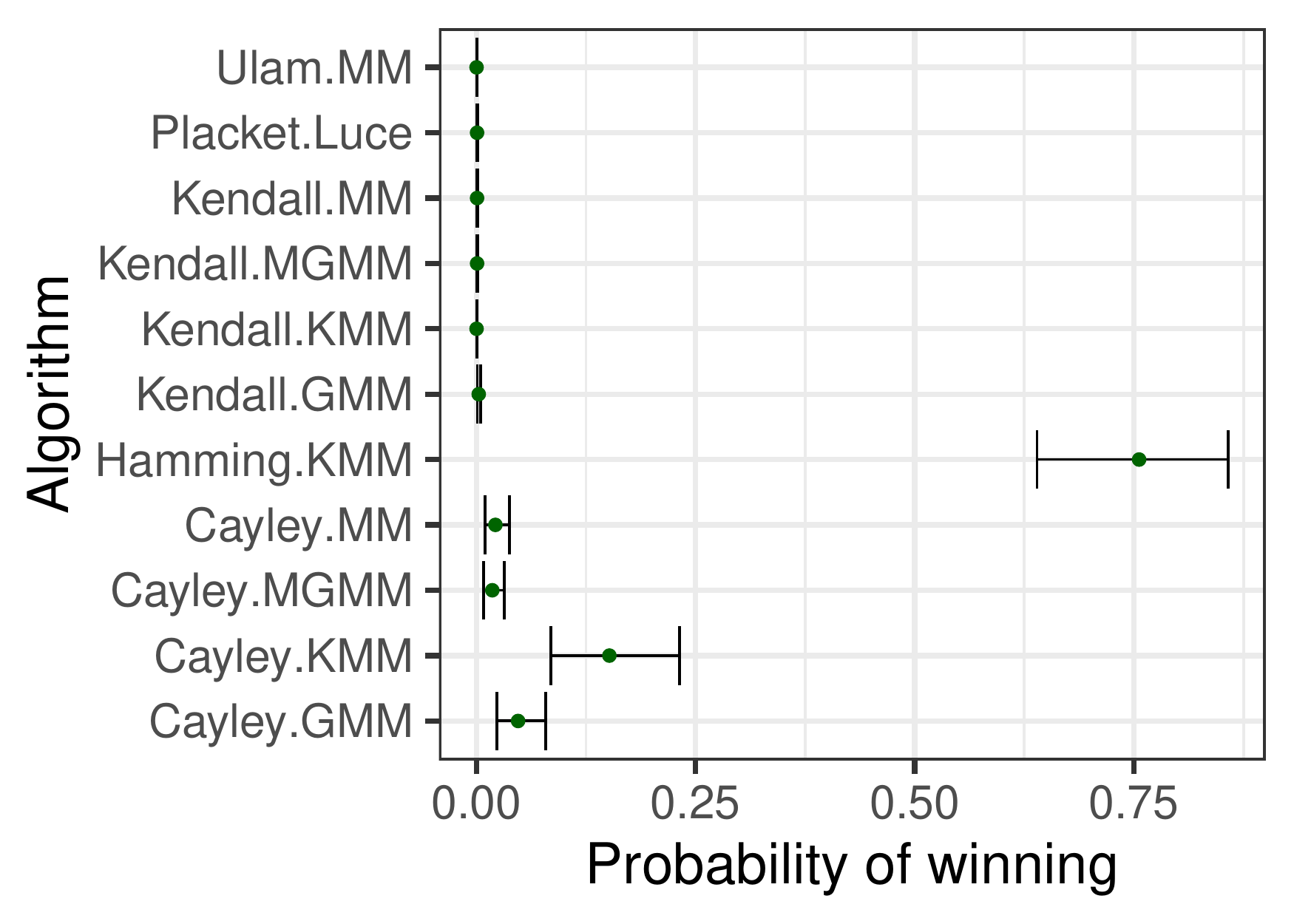}
	\caption{Credible intervals of 90\% and expected value of the estimated posterior probability of each algorithm being the winner among those tested.} 
	\label{fig:native_prob_of_wining_09}
\end{figure}

Figure~\ref{fig:native_prob_of_wining_09} shows the credible interval of $90\%$ of the posterior distribution of the probability of being the best algorithm for the EDAs specific to $\mathcal{S}^n$.
Considering credible intervals of $90\%$, we can say the probability of Hamming KMM being the highest ranked method is higher than $60\%$.
From this analysis, it is clear that Hamming KMM is the algorithm with the highest probability of being the best, followed by Cayley KMM.

\color{black}

To sum up, we observe that KMM EDA obtains a lower ARDP for all the 30 benchmark instances considered.
Therefore, the experimentation suggests that using the proposed model is the best option among the specific EDAs on $\mathcal{S}_n$ for the 30 instances considered in this paper.

\subsection{Experiment 3: Classical EDAs}

	Finally, we compare Hamming KMM EDA to other classical EDAs for the QAP in the literature.
	In the review paper on EDAs in permutation problems \cite{ceberio2012review}, the performance of 13 classical EDAs was studied. 
	Using a null-hypothesis statistical testing, the authors found that there were no statistically significant differences among the best performing six methods for the QAP.
	These six methods are \textit{univariate marginal distribution algorithm} (UMDA) \cite{larranaga2000optimization}, \textit{mutual information maximization for input clustering} (MIMIC) \cite{de1997mimic}, \textit{estimation of Bayesian network algorithm} (EBNA) \cite{bengoetxea2002inexact}, \textit{edge histogram-based sampling algorithm} (EHBSA) \cite{tsutsui2003using} and two variants of \textit{node histogram-based sampling algorithm} (NHBSA) \cite{tsutsui2006node}, namely NHBSA$_{WT}$ and NHBSA$_{WO}$.
	In this experiment, we compare Hamming KMM EDA to these other algorithms.
	Not limited to the previous algorithms, a recent successful EDA, the Random Key EDA \cite{ayodele_rk-eda:_2016, RKIEEE}, was also incorporated to the study.

	The parameters proposed by the EDAs review article~ \cite{ceberio2012review} are used for the methods compared in that article, and we use the parameters proposed by the authors in the RK-EDA~\cite{ayodele_rk-eda:_2016}.
	The ARDP results are shown in Table~\ref{table:othermethods}.

	Hamming KMM EDA obtains the best results in terms of a lower ARPD value in $70\%$ of the considered instances.
	The second most competitive approach is NHBSA$_{WT}$, which outperforms the rest of the methods in $20\%$ of the considered instances.
	In addition to obtaining the lowest ARPD, Hamming KMM EDA is also the most consistent algorithm. 
	For instance, while NHBSA obtains an ARPD over $5\%$ in one instance, the proposed approach obtains lower than $2.5\%$ ARPD in all instances.
	However, for the four \textit{bur26x} instances considered, the \textit{node histogram-based sampling algorithm} (NHBSA$_{WT}$) is able to outperform Hamming KMM EDA.
	These instances have special properties in the distance matrix $D$.
	Specifically, adjacent rows and columns are similar to each other.
	Although Hamming KMM EDA is still the second best approach in these instances, we believe that the Hamming distance is not particularly suited for these instances, as argued in Section \ref{sect:distanceMetrics}.

\begin{table}
	\begin{center}
		\begin{scriptsize}
		\caption{The average ARPD results of Hamming KMM EDA and other EDA approaches. The best performing algorithm is highlighted in bold.}
		\label{table:othermethods}
		\begin{tabular}{lrrrrrrrrrr}

		Instance  & Hamming & UMDA & MIMIC & EBNA &\tiny EHBSA$_{wt}$ & \tiny NHBSA$_{wt}$ & \tiny NHBSA$_{wo}$ & \tiny RK-EDA\\
		&    \myalign{c}{KMM}  & \myalign{c}{\cite{larranaga2000optimization}} & \myalign{c}{\cite{de1997mimic}} & \myalign{c}{\cite{bengoetxea2002inexact}}& \myalign{c}{\cite{tsutsui2003using}}& \myalign{c}{\cite{tsutsui2006node}}& \myalign{c}{\cite{tsutsui2006node}}& \myalign{c}{\cite{ayodele_rk-eda:_2016, RKIEEE}}&  \\
		\hline
		\\
bur26a & 0.105 & 0.323 & 0.281 & 0.311 & 0.442 & \textbf{0.094} & 0.172 & 0.535 \\
bur26b & 0.182 & 0.327 & 0.306 & 0.387 & 0.304 & \textbf{0.095} & 0.238 & 0.475 \\
bur26c & 0.007 & 0.064 & 0.102 & 0.116 & 0.208 & \textbf{0.000} & 0.023 & 0.356 \\
bur26d & 0.007 & 0.063 & 0.146 & 0.073 & 0.021 & \textbf{0.000} & 0.029 & 0.213 \\
nug17 & \textbf{0.179} & 2.760 & 2.200 & 2.673 & 1.386 & 0.202 & 1.247 & 2.991 \\
nug18 & \textbf{0.326} & 2.979 & 3.114 & 2.663 & 2.073 & 0.332 & 1.917 & 2.684 \\
nug20 & \textbf{0.125} & 3.070 & 3.459 & 2.926 & 2.023 & 0.479 & 1.374 & 2.907 \\
nug21 & 0.271 & 2.022 & 2.806 & 1.989 & 3.199 & \textbf{0.254} & 1.214 & 3.868 \\
tai10a & \textbf{0.000} & 2.113 & 3.295 & 2.833 & 1.729 & 0.043 & 1.944 & 5.279 \\
tai10b & \textbf{0.000} & 0.807 & 2.282 & 0.837 & \textbf{0.000} & \textbf{0.000} & 0.461 & 6.617 \\
tai12a & 0.140 & 4.980 & 5.514 & 4.690 & \textbf{0.000} & 0.208 & 4.136 & 7.181 \\
tai12b & \textbf{0.000} & 4.100 & 3.706 & 3.125 & \textbf{0.000} & 0.055 & 1.184 & 10.605 \\
tai15a & \textbf{0.179} & 2.993 & 3.634 & 3.415 & 3.043 & 0.665 & 2.151 & 4.643 \\
tai15b & 0.007 & 0.250 & 0.406 & 0.419 & 0.373 & \textbf{0.000} & 0.163 & 8.724 \\
tai20a & \textbf{0.843} & 4.779 & 5.226 & 4.224 & 4.885 & 2.280 & 3.360 & 7.049 \\
tai20b & \textbf{0.068} & 3.530 & 4.450 & 3.840 & 1.956 & 0.270 & 4.220 & 4.176 \\
tai25a & \textbf{1.265} & 4.387 & 4.700 & 4.297 & 6.160 & 3.630 & 3.325 & 6.510 \\
tai25b & \textbf{0.025} & 2.740 & 3.462 & 2.728 & 1.366 & 0.099 & 0.824 & 9.949 \\
tai30a & \textbf{1.435} & 3.559 & 4.643 & 4.091 & 6.666 & 3.896 & 2.640 & 6.895 \\
tai30b & \textbf{0.189} & 6.502 & 10.143 & 6.621 & 1.332 & 0.765 & 10.801 & 16.502 \\
tai35a & \textbf{1.485} & 4.226 & 4.976 & 4.025 & 7.514 & 4.919 & 2.606 & 7.233 \\
tai35b & \textbf{0.476} & 4.087 & 6.355 & 3.453 & 2.744 & 1.162 & 3.723 & 7.972 \\
tai40a & \textbf{1.762} & 4.038 & 5.246 & 3.771 & 7.959 & 5.292 & 2.748 & 7.814 \\
tai40b & \textbf{1.068} & 5.732 & 8.221 & 5.932 & 4.486 & 2.418 & 5.334 & 9.287 \\
tai60a & \textbf{2.237} & 4.032 & 5.001 & 4.009 & 7.419 & 4.243 & 3.346 & 7.449 \\
tai60b & \textbf{0.493} & 1.188 & 5.309 & 2.558 & 8.177 & 0.836 & 3.379 & 7.347 \\
tai80a & \textbf{2.172} & 3.737 & 4.621 & 3.595 & 7.114 & 4.415 & 3.032 & 7.046 \\
tai80b & \textbf{2.235} & 4.387 & 5.491 & 5.276 & 12.488 & 2.340 & 3.290 & 8.640 \\
tai100a & \textbf{2.190} & 3.460 & 4.227 & 3.321 & 6.737 & 4.519 & 2.792 & 6.636 \\
tai100b & \textbf{1.142} & 2.025 & 4.672 & 2.353 & 11.416 & 1.214 & 2.469 & 5.602 \\

\end{tabular}
\end{scriptsize}
\end{center}
\end{table}

\begin{figure}
	\centering
	\textbf{Probability of winning for EDA approaches non-specific to $\mathcal{S}^n$}\par\medskip
	\includegraphics[width=0.5\linewidth]{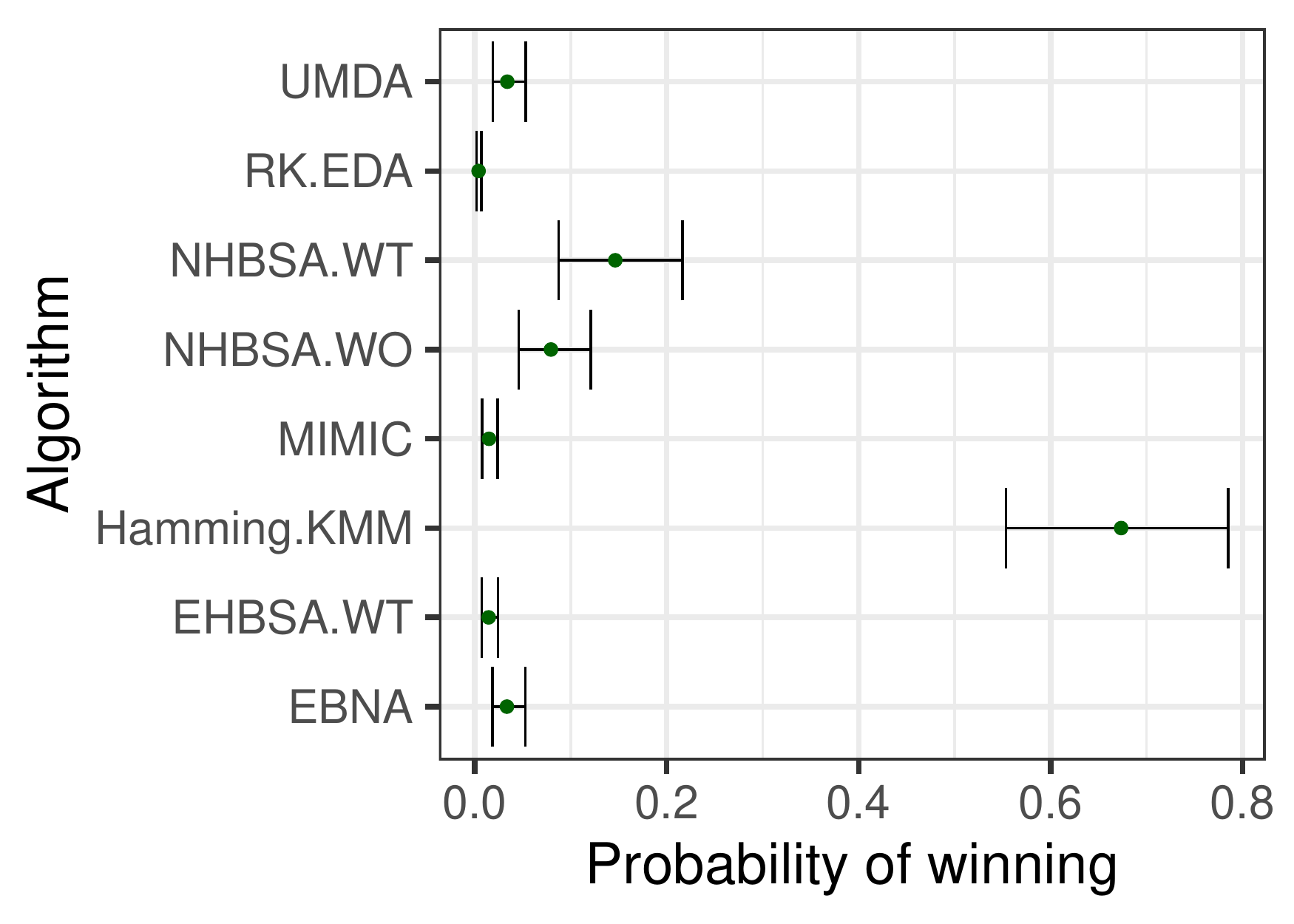}
	\caption{Credible intervals of 90\% and expected value of the estimated posterior probability of each algorithm being the winner among those tested.} 
	\label{fig:non_native_prob_of_wining_09}
\end{figure}

Figure~\ref{fig:non_native_prob_of_wining_09} shows the credible interval of $90\%$ of the posterior distribution of the probability of being the best algorithm for the EDA not specific to $\mathcal{S}^n$.
We can say with high confidence that the probability of Hamming KMM being the highest ranked method is above $50\%$.
In contrast, the probability of the next best performing method, $NHBSA_{wt}$, being the best one is lower than $30\%$.
The rest of the methods have a fairly lower performance, with probability below 0.3 of being the highest ranking methods, considering credible intervals of $90\%$.

Taking into account this analysis, Hamming KMM has a higher chance of being the highest ranked method than the rest of the methods.

	\section{Conclusion \& future work}
	\label{section:conclusion}

	In this paper, we aimed to take a step forward in the development of EDAs for permutation problems.
	We argued that the Hamming distance is suitable for the QAP, as it produces the smoothest objective function transitions when compared to other distance-metrics.
	After analyzing the adequacy of the Hamming distance for the QAP, we proposed an algorithm that implements a Hamming-based Kernels of Mallows Models (KMMs) EDA.
	In order to analyze the performance of the proposed approach, we compare it to other non-hybrid EDAs presented in the literature.
	The conducted experimentation showed that, for the QAP, (i) Hamming KMM EDA performs better than other classical EDAs, (ii) it also performs better than other MM approaches in the literature, and (iii) the use of Kernels on a Hamming-based MM is the key to the successful performance of the algorithm.
	Specifically, Hamming KMM EDA is able to outperform the rest of the methods in $56.7\%$ of the studied instances.
	Not only that, but Hamming KMM EDA is also more stable than the competitors in terms of maximum ARDP, with an ARDP value of less than $2.5\%$ in the most difficult instance for this algorithm.

	The incorporation of Hamming-based KMMs to the EDA framework in a competitive manner opens new research directions worth considering. 
	For instance, this method could potentially be applied to other permutation problems, and even in non-permutation based combinatorial problems, if the solution space of the problem can be encoded by vectors. 
	Because the Hamming distance measures the mismatches, regardless of the order, we believe that this method could be especially successful in combinatorial problems where the order of the elements in the vector is not as relevant as the absolute position of the items, such as the graph-partitioning problem \cite{bulucc2016recent}.

	\section*{Acknowledgments}
	This work was partially supported by Basque Government (Elkartek program), by the Spanish Government (BERC program 2018-2021, the project TIN2016-78365-R, the project PID2019-106453GA-I00, BCAM Severo Ochoa excellence accreditation SVP-2014-068574 and SEV-2013-0323), and by  AEI/FEDER, UE (through the project TIN2017-82626-R).

	\section*{References}

	\bibliographystyle{abbrv}
	\bibliography{sample}

\end{document}